# Advancing Improvisation in Human-Robot Construction Collaboration: Taxonomy and Research Roadmap

David Wireko Atibila[1], Vineet R. Kamat[2], Carol C. Menassa[3]

[1]PhD Student, Department of Civil and Environmental Engineering, University of Michigan, University of Michigan, 2350 Hayward Street, 2340 G.G. Brown Building, Ann Arbor, MI 48109, USA, E-mail: datibila@umich.edu

[2] Professor, Department of Civil and Environmental Engineering, University of Michigan, University of Michigan, 2350 Hayward Street, 2340 G.G. Brown Building, Ann Arbor, MI 48109, USA, E-mail: vkamat@umich.edu

[3] Professor, Department of Civil and Environmental Engineering, University of Michigan, University of Michigan, 2350 Hayward Street, 2340 G.G. Brown Building, Ann Arbor, MI 48109, USA, E-mail: menassa@umich.edu

**Abstract**

The construction industry faces productivity stagnation, skilled labor shortages, and safety concerns. While robotic automation offers solutions, construction robots struggle to adapt to unstructured, dynamic sites. Central to this is improvisation, adapting to unexpected situations through creative problem-solving, which remains predominantly human. In construction's unpredictable environments, collaborative human-robot improvisation is essential for workflow continuity. This research develops a six-level taxonomy classifying human-robot collaboration (HRC) based on improvisation capabilities. Through systematic review of 214 articles (2010-2025), we categorize construction robotics across: Manual Work (Level 0), Human-Controlled Execution (Level 1), Adaptive Manipulation (Level 2), Imitation Learning (Level 3), Human-in-Loop BIM Workflow (Level 4), Cloud-Based Knowledge Integration (Level 5), and True Collaborative Improvisation (Level 6). Analysis reveals current research concentrates at lower levels, with critical gaps in experiential learning and limited progression toward collaborative improvisation. A five-dimensional radar framework illustrates progressive evolution of Planning, Cognitive Role, Physical Execution, Learning Capability, and Improvisation, demonstrating how complementary human-robot capabilities create team performance exceeding individual contributions. The research identifies three fundamental barriers: technical limitations in grounding and dialogic reasoning, conceptual gaps between human improvisation and robotics research, and methodological challenges. We recommend future research emphasizing improved human-robot communication via Augmented/Virtual Reality interfaces, large language model integration, and cloud-based knowledge systems to advance toward true collaborative improvisation.

**Introduction**

The construction industry faces persistent challenges of productivity stagnation, skilled labor shortages, and safety concerns. The United States (US) construction industry recorded 1,075 workplace fatalities in 2023, representing the highest number among all industry sectors and continuing a trend that has persisted since 2011 (US BLS, 2024). Compounding these safety challenges, only 25% of construction projects finish within 10% of their original deadline, underscoring the urgent need for innovative approaches addressing both safety and performance simultaneously (KPMG, 2015; Shah, 2016).

Automation and robotics have emerged as promising solutions for relieving human workers from hazardous conditions while improving project efficiency and quality (Bock, 2007; Saidi et al., 2008). Recent advances in hardware, software, and machine learning have significantly progressed construction robotics capabilities, leading to the emergence of collaborative robot teams (cobots) designed to work alongside human workers (Balaguer and Abderrahim, 2008; You et al., 2018). However, construction's unstructured, dynamic environments present unique challenges distinguishing it from manufacturing applications. Traditional preprogrammed automation systems execute predetermined tasks with high precision but limited flexibility (Liang et al., 2020), struggling when confronted with unexpected conditions characteristic of construction sites and requiring extensive human supervision (Pradhananga et al., 2021; Wang et al., 2021). This gap between robotics promise and practical implementation has given rise to human-robot collaboration (HRC) paradigms, where human workers and robots share tasks and coordinate activities to accomplish construction work more effectively than either could achieve independently (Bock, 2015; Chu et al., 2013; Liang et al., 2021; Olukanni et al., 2026).

Construction fundamentally operates as a team-based endeavor where collective performance emerges from coordinated integration of complementary skills rather than individual excellence alone, analogous to team sports where diverse player capabilities combine synergistically to achieve outcomes unattainable by any single participant (Salas et al., 2005; Marks et al., 2001). Just as successful football teams leverage diverse player capabilities, combining the quarterback's strategic vision with receivers' speed and linemen's strength, construction projects demand synergistic coordination among trades, roles, and increasingly, human-robot partnerships (Raiden et al., 2004; Dainty et al., 2004). This team perspective reframes automation not as replacement but as augmentation, where human-robot partnerships leverage complementary strengths which is human contextual wisdom with robotic computational power to expand the collective capability envelope beyond what either agent could accomplish independently (Steinfeld et al., 2006; Tsarouchi et al., 2016).

Effective HRC in construction depends on how well humans and robots complement each other across multiple capability dimensions: strategic planning and task sequencing, cognitive reasoning for real-time decision-making, physical execution of construction tasks, and continuous learning to adapt to new situations (Goodrich and Schultz, 2007; Bauer et al., 2008). Optimal collaboration emerges when human contextual understanding, creativity, and adaptive reasoning combine with robotic precision, consistency, and computational power to achieve outcomes neither agent could accomplish independently (Steinfeld et al., 2006; Tsarouchi et al., 2016). While much attention has focused on physical capabilities and basic coordination, one critical skill essential for effective task completion has received limited attention in construction HRC literature.

Central to the success of human-robot collaboration in construction is the concept of improvisation, the ability to adapt and respond to unexpected situations through creative problem-solving and real-time decision-making (Hamzeh et al., 2019). Improvisation in construction has been recognized as a deliberate, spontaneous, and rational decision-making process that helps address emerging issues or unplanned work when traditional planning approaches fall short. Research by Menches and Chen (2013) demonstrated that construction workers frequently rely on improvisational skills to continue working when disruptions occur, with every disruption requiring some form of adaptive response. The dyadic nature of improvisation in construction reveals that improvisational decision-making typically involves collaborative teams, where forepersons often improvise new plans while crew members execute these improvised solutions (Menches et al., 2013).

The concept of collaborative improvisation in human-robot teams represents a significant departure from conventional automation paradigms. Rather than replacing human workers with predetermined robotic processes, this approach envisions partnerships where humans and robots jointly engage in creative problem-solving. Recent advances in artificial intelligence, machine learning, and human-robot interaction have begun opening new possibilities for more adaptive and collaborative robotic systems (Ardiny et al., 2015; Billard and Kragic, 2019; Park et al., 2024; Ravichandar et al., 2020). Large Language Models (LLMs) integrated with robotic control systems demonstrate potential for enabling more intuitive communication and flexible task execution (Huang et al., 2022), while virtual and augmented reality technologies offer promising platforms for developing and testing collaborative improvisation capabilities (Park et al., 2024; Wang et al., 2018; Williams et al., 2019).

Current research in construction robotics has primarily focused on structured approaches to human-robot interaction, ranging from preprogrammed systems to adaptive manipulation capabilities (Liang et al., 2021; Zeng et al., 2019; Park et al., 2024). However, there exists a significant gap between the current capabilities of construction robots, which largely operate under human supervision with minimal autonomous adaptation, and the dynamic requirements of construction environments that frequently demand collaborative improvisation when unexpected situations arise. Current limitations preventing collaborative improvisation in construction robotics include the lack of systems capable of autonomous creative reasoning, inadequate frameworks for real-time collaborative decision-making between humans and robots, and insufficient integration of human knowledge and experience into robotic decision-making processes.

This challenge is particularly acute in construction due to the industry's inherent uncertainty and variability. Unlike manufacturing environments where robots execute repetitive tasks in controlled conditions, construction sites present constantly changing conditions, unique project requirements, and unforeseen obstacles requiring adaptive responses (Abdelhamid et al., 2009). Developing improvisational capabilities in construction robotics must account for the complex interplay between human creativity, experience, and intuition, and robotic precision, consistency, and computational power (Han and Parascho, 2022; Thörn et al., 2020; Abdelhamid et al., 2009). The challenge lies in developing systems that complement and enhance human capabilities rather than simply replacing them, creating truly collaborative teams capable of joint creative problem-solving. The potential benefits of achieving true collaborative improvisation in human-robot construction teams are substantial. Such capabilities could enable construction projects to respond more effectively to unexpected conditions, reduce delays and cost overruns associated with planning failures (Han and Parascho, 2022; Thörn et al., 2020), improve safety by providing adaptive responses to hazardous situations (Awolusi et al., 2018; Cheng et al., 2013), and enhance overall project quality through the combination of human creativity and robotic precision (Delgado et al., 2020).

Given improvisation's critical importance in construction work and growing robotic capabilities, comprehensive research is urgently needed to systematically examine the progression from current HRC approaches toward true collaborative improvisation capabilities. This research must address fundamental questions: How can human-robot collaboration in construction be classified based on improvisation capabilities? What technical and conceptual frameworks are required to enable collaborative improvisation? What current limitations prevent such capabilities? How do construction workers currently handle improvisation and what role could robots play in these processes? Developing a systematic classification framework for understanding the progression toward collaborative improvisation in human-robot construction teams is essential for guiding future research and development efforts, providing clear pathways for advancing from current capabilities toward true collaborative improvisation while identifying specific technical requirements, research gaps, and development priorities.

**Research Objectives and Methodology**

This research aims to systematically examine the evolution of improvisation capabilities in human-robot collaboration for construction by developing a comprehensive classification framework. The specific objectives are to: (1) adapt existing taxonomies from literature to categorize construction human robot collaboration research based on improvisation capabilities and the corresponding distribution of responsibilities between human workers and robotic systems; (2) systematically review and classify existing construction robotics research within this taxonomy framework; (3) identify current technological and conceptual limitations preventing the achievement of true collaborative improvisation; and (4) recommend future research directions necessary for advancing construction human-robot teams toward higher levels of improvisational autonomy.

The research methodology consisted of three primary phases: literature collection, classification framework development, and systematic analysis as illustrated in Figure 1. In the literature collection phase, relevant articles were identified using Google Scholar, IEEE Xplore, Scopus, and ScienceDirect databases. The search strategy employed Boolean combinations including ("construction robot" OR "construction automation" OR "building robotics") AND ("human-robot collaboration" OR "improvisation" OR "adaptation" OR "imitation learning" OR "human-in-loop" OR "BIM"). For databases using keyword-based searches, individual keywords were employed including "construction robotics," "construction automation," "building robotics," "human-robot collaboration," "construction improvisation," "adaptive manipulation," "imitation learning," "BIM robotics," and "collaborative construction." Additional searches targeted specific journals including *Journal of Computing in Civil Engineering*, *Journal of Construction Engineering and Management*, *Automation in Construction*, and *Construction Robotics*, as well as proceedings from the International Symposium on Automation and Robotics in Construction (ISARC) and Construction Research Congress (CRC). The search timeframe focused on publications from 2010 to 2025, justified by the emergence of modern human-robot collaboration research during this period. Duplicate articles were identified and removed during the screening process.

Manual screening filtered articles based on relevance criteria including robot systems, automated equipment, control methods, and motion-planning algorithms. This process yielded 214 relevant articles for detailed analysis. To specifically emphasize improvisation capabilities in construction contexts, a six-level taxonomy was established by adapting Liang et al.'s (2021) five-level human-robot collaboration framework which itself built upon the Level of Robot Autonomy (LoRA) framework for general human-robot interaction (Beer et al, 2014). The framework was expanded to include Manual Work as a baseline level and refined to distinguish between consultative autonomy (Level 5) and true co-creative improvisation (Level 6), addressing the unique challenges of quasi-repetitive tasks and unstructured construction environments. Each level was defined by the progressive distribution of sensing, planning, and acting responsibilities between humans and robots, with particular emphasis on the evolution from zero improvisation to collaborative creative problem-solving capabilities.

The systematic analysis phase involved categorizing selected articles into the six taxonomy levels, assessing current capabilities and limitations at each level, and conducting gap analysis to identify research frontiers. A radar diagram visualization framework was developed to illustrate the progressive evolution of five critical capability dimensions, and these are Planning, Cognitive Role, Physical Execution, Learning Capability, and Improvisation across the six collaboration levels, providing a comprehensive representation of human-robot capability distribution throughout the maturity spectrum. Additionally, spoke-map representations were constructed to locate representative research contributions within each level, based on their emphasis on cognitive, physical, or adaptive problem-solving characteristics.

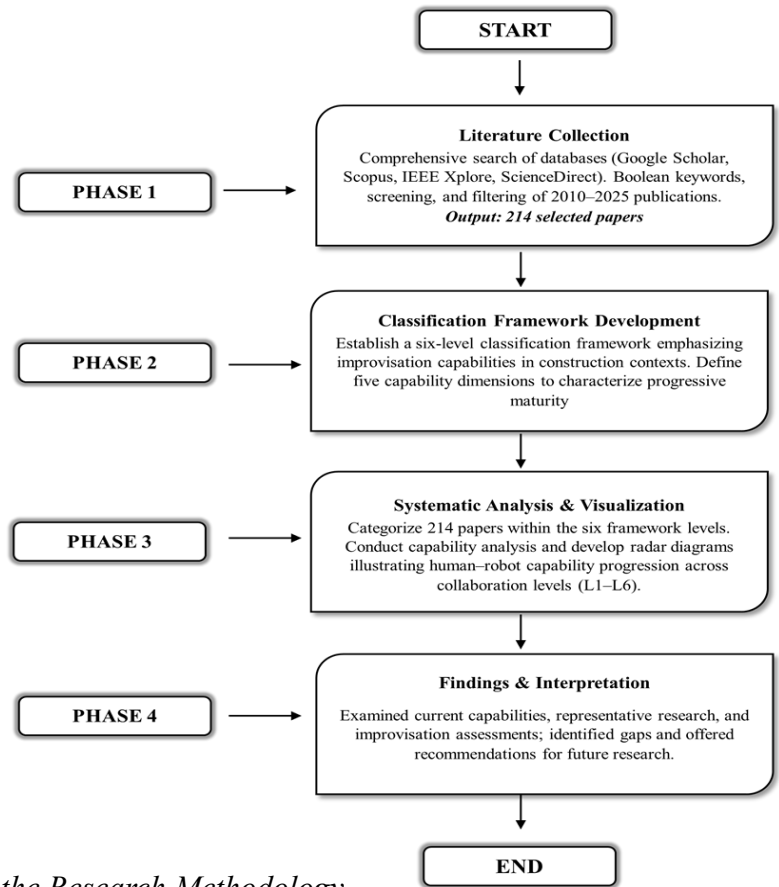

*Fig 1: Overview of the Research Methodology*

**Classification Framework – Rationale and Overview**

The progression toward collaborative improvisation in Human-Robot Collaboration (HRC) can be understood through the systematic evolution of how planning, decision-making, physical execution, learning, and improvisational capabilities are distributed between humans and robots. The foundational framework for understanding robot capabilities in construction centered on three core dimensions: planning (high-level task organization), cognitive (task-level decision-making), and physical (direct manipulation and execution), initially developed for general robotics and recognized as fundamental to construction automation (Beer et al., 2014; Everett and Slocum, 1994; Saidi et al., 2016).

The dynamic and quasi-repetitive nature of construction tasks where similar operations must adapt to varying conditions, motivated investigating learning as a fourth dimension, enabling robots to generalize from demonstrations rather than requiring explicit programming for every variation (Billard et al., 2016; Liang et al., 2020; Osa et al., 2018). Building on this progression, the current research proposes improvisation as a fifth essential dimension. While learning handles variations within known task families, improvisation addresses unlearned situations where neither prior demonstrations nor accumulated experience provide direct guidance which is critical in construction's unpredictable environments where unexpected obstacles and conflicts routinely occur (Hamzeh et al., 2019; Menches and Chen, 2013). Critically, this framework characterizes capabilities not as individual robot attributes, but as distributed team competencies shared between human workers and robotic systems. This perspective draws from research on high-performing teams where optimal performance emerges from complementary rather than identical capabilities among members (Marks et al., 2001; Salas et al., 2005).

To represent this progression, this study adopts the five-dimensional capability framework to characterize core competencies required for construction task execution and provide a basis for visualizing how human and robot contributions change across six collaboration maturity levels (L1-L6). The framework also forms the analytical basis for assessing the evolving division of responsibilities as collaboration advances from human-dominant execution toward shared, adaptive, and ultimately co-creative improvisation. The sections below define the five capability dimensions, and Figure 2 introduces the radar diagram representation used to compare capability distributions across collaboration levels.

**Five-Dimensional Capability Framework: Definitions**

The classification of human-robot collaboration in construction requires a comprehensive understanding of the capability dimensions that define both human and robotic contributions to construction tasks. This framework establishes five critical dimensions that collectively characterize the competencies necessary for effective construction task execution in increasingly complex and uncertain environments.

*Planning / Organization*

Constitutes the highest level of project management functions, encompassing task sequencing, resource scheduling, crew coordination, and resolution of inter-trade conflicts (Saidi et al., 2008; Baek et al., 2026). For human workers, planning manifests as anticipating project needs, coordinating multiple trades, and dynamically resequencing activities based on contextual factors. For robotic systems, planning represents the degree of autonomous task organization, from executing fixed sequences (low) to autonomously generating multi-step plans and proposing alternatives based on real-time conditions (high). Capability ranges from simple single-trade task sequencing (low), through single-discipline coordination (moderate), to comprehensive multi-trade coordination with conflict resolution and resource optimization (high) (Everett and Slocum, 1994).

*Cognitive Role*

Addresses task-level decision-making and reasoning capabilities essential for immediate task execution, encompassing real-time path planning, obstacle detection and avoidance, adaptive manipulation, sensor data interpretation, and elemental motion control (Beer et al., 2014; Feng et al., 2015; Lundeen et al., 2017). For human workers, cognitive capability reflects the ability to perceive environments, recognize patterns, make rapid tactical decisions, and adjust actions based on immediate sensory feedback. For robotic systems, cognitive capability represents the sophistication of perception-action loops, from simple preprogrammed responses (low) to complex scene understanding and context-aware decision-making (high). Capability progresses from pre-programmed motions with no adaptive response (low), through sensor-driven adjustments (moderate), to sophisticated real-time reasoning with contextual understanding and predictive adjustments (high) (Liang et al., 2021; Baek et al., 2026).

*Physical Task Execution*

Represents direct manipulation, material handling, tool operation, and physical labor dimensions of construction activities, encompassing both gross motor functions and fine motor skills (Saidi et al., 2008). For human workers, physical capability includes strength, endurance, dexterity, and kinesthetic awareness, excelling in adaptability across diverse tasks. For robotic systems, physical capability reflects mechanical design, actuation power, and force control sophistication, providing superhuman strength and precision but traditionally lacking human adaptability. Capability extends from minimal physical involvement characterized by guidance or supervisory roles, through moderate force application and basic dexterity, to sustained heavy lifting, repetitive precision work, or exceptional dexterity at the high end (Bock, 2007).

*Learning Capability*

Characterizes the agent's capacity for continuous improvement and generalization through various modalities including experience, demonstration, formal instruction, or data synthesis (Beer et al., 2014; Liang et al., 2020). For human workers, learning manifests as acquiring new skills through observation, practice, and experience, transferring knowledge across contexts, and continuously refining techniques. For robotic systems, learning capability represents the sophistication of algorithms enabling skill acquisition from zero learning capacity beyond initial programming (low) to advanced machine learning enabling pattern extraction, experience accumulation, and autonomous improvement (high). Capability progresses from complete inability (low) to learn beyond programming, through moderate generalization from demonstrations within known contexts, to high-level continuous learning from diverse experiences with multi-source knowledge integration and adaptation to novel situations (Osa et al., 2018; Rozo et al., 2016).

*Improvisation and Adaptability*

Represents the capacity to generate novel solutions for unexpected situations through creative problem-solving, consultation, knowledge synthesis, or autonomous reasoning (Hamzeh et al., 2019). For human workers, improvisation draws upon experiential knowledge, analogical reasoning, and creative problem-solving, substituting methods, consulting peers, and generating solutions never explicitly taught (Menches and Chen, 2013). For robotic systems, improvisation capability represents the degree to which systems can autonomously respond to situations outside their training distribution from complete inability (low) to sophisticated reasoning systems that query knowledge bases and generate creative solutions grounded in physical constraints (high). Low capability reflects inability to respond to unexpected conditions, moderate enables adaptation within learned parameters with guidance-seeking, while high supports generation of creative solutions independently and synthesis of knowledge from multiple sources (Liang et al., 2021; Menches and Chen, 2013).

**Human Capability Variation**

The interaction among these five dimensions creates a comprehensive capability profile for both human workers and robotic systems. The framework acknowledges that human capabilities vary significantly based on experience level, training background, and trade specialization, while robotic capabilities progress systematically through technological advancement (Loosemore and Waters, 2004; Abraham et al., 2025). The baseline human capability profile in this framework represents an average experienced construction worker with approximately 3-7 years of field experience, recognizing that novice workers exhibit substantially lower capabilities in improvisation and cognitive dimensions, while expert workers demonstrate significantly higher capabilities across all dimensions (Schubert et al., 2013; Raiden et al., 2008). Trade specialization further differentiates capability profiles, with electricians and plumbers developing distinct cognitive and troubleshooting skills compared to equipment operators or structural trades workers (Dainty et al., 2004; Zuo and Zillante, 2005). This variability underscores that effective human-robot collaboration requires not merely additive capability combination, but rather a multiplicative effect wherein human contextual wisdom guides robotic computational power toward optimal solutions (Liang et al., 2021; Saidi et al., 2016).

**Radar Diagram Representation**

The radar diagram representation (Figure 2) provides a visual framework for understanding the progressive evolution of capability distribution between human workers and robotic systems across the six levels of Human-Robot Collaboration (HRC) maturity in construction. This visualization approach was adapted from the taxonomy proposed by Liang et al. (2021) for human-robot collaboration in construction, which itself built upon the Level of Robot Autonomy (LoRA) framework for general human-robot interaction (Beer et al, 2014). The framework was modified to explicitly emphasize the improvisation capabilities central to construction work and the complementary nature of human-robot teaming. Each diagram maps five critical dimensions, Planning, Cognitive Role, Physical Execution, Learning Capability, and Improvisation, that collectively define the competencies required for effective construction task execution in increasingly complex and uncertain environments. The orange solid line represents human worker capabilities, while the blue dashed line depicts robot capabilities.

Critically, the team's total capability at any given level is represented by the superset (or bounding box) encompassing both profiles. This representation acknowledges that effective HRC does not require identical capabilities from both agents; rather, it requires complementary strengths that, when combined, exceed what either agent could achieve independently. As the collaboration matures from Level 1 (Human-Controlled Execution) to Level 6 (True Collaborative Improvisation), the robot's capabilities systematically expand often "punching through" the human capability boundaries on specific dimensions thereby enlarging the overall team capability envelope. However, this expansion is intentionally bounded by design principles that preserve essential human roles, particularly in high-level planning and contextual learning, ensuring that robots augment rather than replace human expertise.

Figures 2a through 2f (presented in subsequent sections alongside detailed analyses) apply this radar framework across all six collaboration levels. Each figure reveals specific capability transitions: robots progressively gain autonomy in physical execution, cognitive reasoning, and learning, while humans evolve from direct operators to supervisors and ultimately co-creative partners. The detailed spoke-map representations (Figures 3-8) further locate representative research contributions within each level based on their emphasis on cognitive, physical, or adaptive problem-solving characteristics. Together, these visualizations form the analytical foundation for systematically examining current research capabilities and identifying critical gaps toward achieving true collaborative improvisation in construction robotics.

**Analysis of Collaboration Levels: Progressive Maturity Path**

The six collaboration levels represent progressive maturity stages characterized by evolving capability distributions between humans and robots across the five-dimensional framework, illustrating systematic transitions from human-dominant execution toward complementary partnerships while preserving essential human roles in contextual reasoning and creative problem-solving.

**Level 1: Human-Controlled Execution**

*Characteristics and Team Performance*

L1 functions as a tool-like paradigm where robots serve as obedient executors under complete human control. Humans maintain High capability across planning, cognitive, and learning dimensions, directing robot actions through teleoperation or preprogrammed instructions. Robots contribute High physical capability (70%), relieving humans from strenuous labor, but possess zero autonomous decision-making, learning, or adaptation. Human improvisation remains Moderate (50%), representing average-experienced workers who solve routine problems independently but require consultation for complex disruptions. Robot improvisation is absent (0%), causing system halt at every unexpected condition (Figure 2a). The team achieves High planning, cognitive, learning, and physical capabilities through complementary contributions, but only Moderate improvisation, limited by human baseline. This configuration suits highly predictable, repetitive tasks in controlled environments, such as block stacking, repetitive welding, automated material handling. The critical bottleneck is that humans must micromanage every robot motion, preventing improvisation beyond individual human capacity. L2 advances this by developing sensory feedback and adaptive manipulation enabling minor autonomous adjustments (Liang et al., 2021; Saidi et al., 2016).

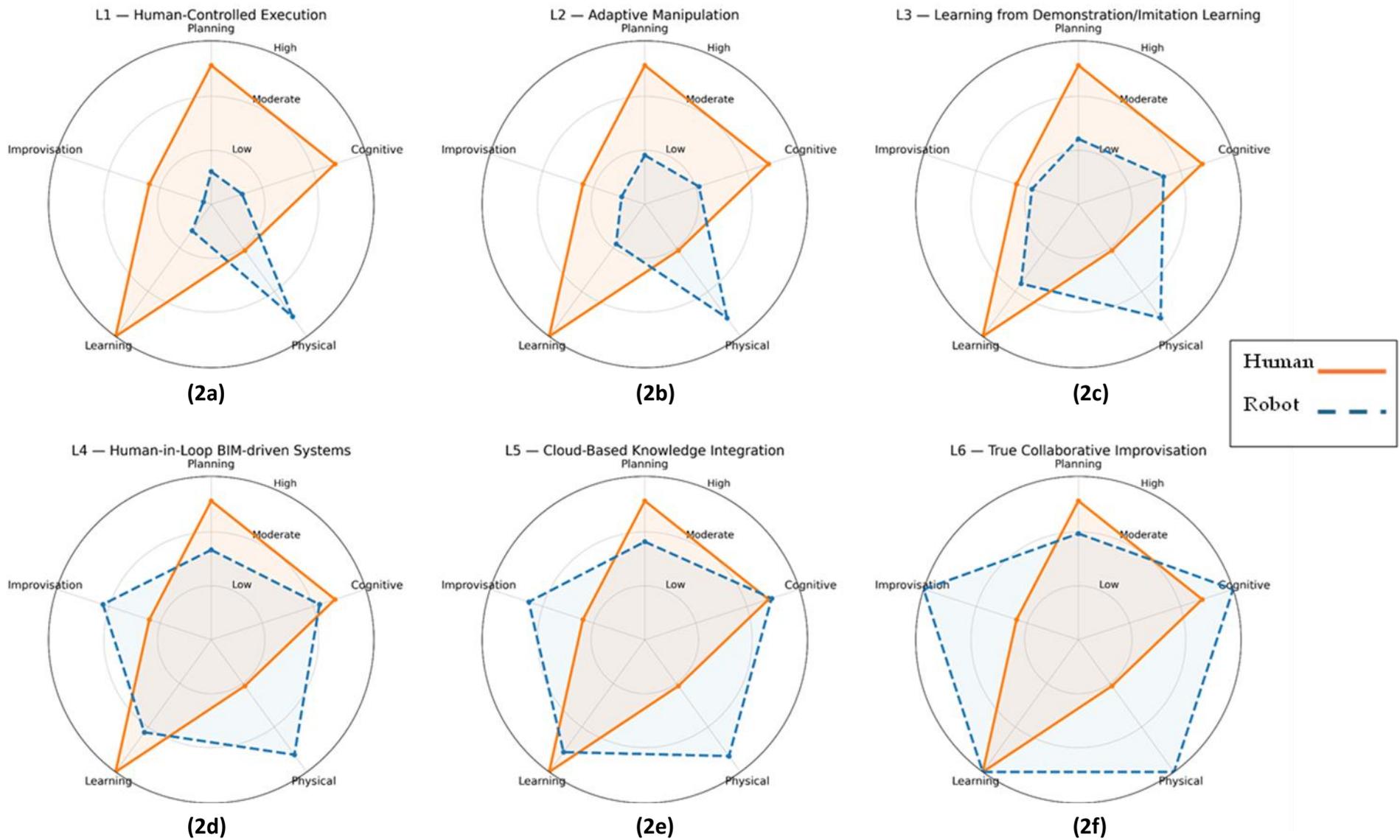

*Fig 2:* Radar Diagram of Human and Robot Capability Distribution Across Collaboration Levels (L1-L6). Human capabilities are shown at middle-range baseline representing experienced workers, robot capabilities progress from low to high across levels.

**Level 2: Adaptive Manipulation**

*Capabilities and Team Performance:*

L2 introduces robot autonomy through adaptive manipulation, robot cognitive capability advances to Moderate (40%), enabling sensor-driven micro-adjustments for position correction, force modulation, and collision avoidance. Human cognitive load decreases to 70% as they transition from continuous commanding to monitoring and intervening only when adjustments exceed tolerance. Robot physical capability increases to 80%, with minimal learning (10%) and improvisation (15%) constrained to pre-specified ranges (Figure 2b). Masonry robots adjust brick placement by millimeters based on laser feedback but cannot alter laying patterns for poor mortar consistency.

The team achieves improved efficiency on repetitive precision tasks with minor variations. Robots "punch through" human boundaries on Physical capability, demonstrating complementary specialization. Applications include automated bricklaying with laser-guided corrections, drilling operations adjusting for material hardness, concrete finishing with pressure modulation, and steel assembly compensating for minor misalignments (Lundeen et al., 2019; Feng et al., 2015; Gambao et al., 2000). L2's limitation is robots adapt only within narrow predefined parameters through reactive feedback rather than anticipatory reasoning (Fitzgerald et al., 2021). L3 advances through learning from demonstration, enabling robots to build skill repertoires for novel situations (Osa et al., 2018; Liang et al., 2020).

**Level 3: Learning from Demonstration (LfD)**

*Characteristics and Team Performance:*

L3 shifts from programming to teaching that is humans demonstrating desired behaviors while robots observe, encode, and generalize into executable policies. Robot learning capability increases to Moderate (45%), acquiring skills from kinesthetic guidance, teleoperation, or video observation. Robot cognitive capability (55%) approaches human levels (60%), enabling task-level decisions including motion primitive selection, workspace adjustment, and sub-task sequencing. Robot improvisation develops modestly (25%), constrained to interpolating between learned examples (Figure 2c).

The team efficiently handles quasi-repetitive tasks, activities sharing common structure but varying in parameters. Examples include curtain wall installation with varying panel sizes, rebar tying adapting grip pressure for different diameters, and formwork assembly reproducing strategies with geometric adjustments (Liang et al., 2019, 2020). Robots approach human parity on cognitive tasks, signaling complementary partnership. The defining L2 to L3 progression is robot transition from Low to Moderate learning through experience-driven skill acquisition. L3's limitation is learning remains experiential and isolated, preventing access to broader knowledge contexts (Kim et al., 2022; Viano et al., 2022). L4 advances by integrating BIM and Digital Twins, providing structured knowledge about design intent and project context (Dai et al., 2024; Vieira et al., 2023).

**Level 4: Human-in-the-Loop BIM**

*Characteristics and Team Performance:*

L4 enables robots to integrate BIM/Digital Twin data for autonomous task execution updates. Robot planning capability reaches Moderate (40%), enabling task sequencing, discrepancy detection between as-designed and as-built conditions, and corrective action proposals, though humans retain high-level coordination authority. Robot cognitive capability (65%, Moderate-High) punches through human baseline for the first time, exceeding average workers in geometric calculations and sensor fusion. Robot improvisation reaches Moderate (45%), approaching human baselineadapting plans based on digital model updates. Human improvisation increases to 55% as supervisors must validate robot proposals and resolve BIM-reality conflicts (Figure 2d).

The team operates in human-in-the-loop autonomy where robot cognitive contributions first exceed human contributions on certain tasks, demonstrating true complementarity. Applications include automated quality inspection flagging BIM-scan discrepancies, prefabricated component installation proposing alternative sequences when site conditions differ from design, and MEP coordination detecting clashes and suggesting corrective routing (Dai et al., 2024; Kavaliauskas et al., 2022). The team achieves High capabilities across all dimensions with Moderate improvisation. The defining L3 to L4 progression is robot achieving Moderate planning and improvisation plus Moderate-High cognitive capability exceeding human baseline in computational reasoning. L4's limitation is robot knowledge remains bounded by specific BIM models and isolated experience, restricting improvisation to single-project scope (Ramasubramanian et al., 2022). L5 advances through cloud robotics enabling access to collective expertise and historical precedents (Kehoe et al., 2015; Firoozi et al., 2024).

**Level 5: Cloud-Based Knowledge (Horizontal/Vertical Consultation)**

*Characteristics and Team Performance:*

L5 represents the current research frontier where robots leverage cloud-connected knowledge bases for encoded expertise and collective learning from distributed fleets. Robot improvisation (High) punches through human baseline (55%) for the first time, enabled by instantaneous access to knowledge requiring years of human experience accumulation. Robot planning approaches 50% (48%), generating multiple alternatives using cloud-sourced best practices while humans retain final authority. Robot cognitive (High) and learning (High) capabilities enable reasoning through complex scenarios by querying cloud networks for analogous cases. Humans provide strategic validation, contextual judgment transcending algorithmic optimization, and handling novel situations without knowledge base precedent (Figure 2e).

The team operates as knowledge-augmented partnership where, for the first time, High improvisation is achieved through robot contributions, enabling less-experienced workers (Moderate improvisation) to achieve expert-level problem-solving. Scenarios include novice workers executing complex installations with robots accessing worldwide successful strategies, excavation operations querying soil databases for alternative digging patterns, and adaptive formwork consulting case libraries for pour sequence adjustments (Kehoe et al., 2015; Firoozi et al., 2024; You et al., 2023). The team achieves High capabilities across all dimensions. The defining L4 to L5 progression is robot achieving High cognitive, learning, and improvisation capabilities with improvisation first exceeding human baseline, transitioning from human-dependent to AI-augmented creative reasoning. L5 remains fundamentally consultative and sequential rather than collaborative, following query-response patterns rather than joint creative reasoning (Yoshikawa et al., 2023). L6 requires AR/VR interfaces for shared mental models, LLMs for real-time collaborative ideation, and dynamic autonomy frameworks, transitioning from intelligent assistant to creative partner (Chen et al., 2024; Lakhnati et al., 2024).

**Level 6: True Collaborative Improvisation**

*Characteristics and Team Performance:*

L6 represents the aspirational pinnacle, true partnership where both agents engage in continuous, co-creative problem framing, exploration, and execution in real time. This represents fundamental transition beyond knowledge-assisted inference toward co-creative cognition, integrating generative models, on-site digital twins, and knowledge infrastructures with embodied actuators for emergent solution generation amid open-world uncertainty (Hautala and Jauhiainen, 2022; Chen et al., 2024).

L6 achieves near-parity through intentional asymmetries preserving essential human roles. Planning shows robots at moderate ceiling (50%) while humans maintain high capability for ethical strategic decisions. Cognitive dimension shows high robot capability for computational reasoning while humans focus on contextual interpretation. Physical execution demonstrates high robot capability with minimal human contribution. Humans retain high learning capability for contextual wisdom acquisition transcending data-driven machine learning. Improvisation demonstrates high robot capability (85%) through advanced AI including LLMs and generative reasoning, while humans provide moderate capability for contextual constraints, ethical judgment, and final validation (Figure 2f).

The team operates as truly collaborative cognitive system where both agents contribute to problem framing, solution generation, and execution refinement in real-time, with team capability approaching full pentagon coverage across all dimensions. This enables handling unprecedented problems in unstructured environments which includes emergency structural repairs requiring creative material adaptation, complex renovation in historically protected buildings, and disaster response construction requiring collaborative solution generation for unstable conditions with limited information (Han et al., 2021; Onososen and Musonda, 2022). The defining L5 to L6 progression is robot achieving maximum planning (50%) and peak improvisation (85%) while maintaining High cognitive and learning, representing joint creative problem-solving through sustained dialogic reasoning rather than sequential query-response.

**Current State Analysis: Systematic Review of Improvisation in HRC Levels**

Having established the theoretical framework, the section now examines what has actually been achieved across the six collaboration levels through detailed analysis of representative research. Each level is assessed through current capabilities, representative research contributions positioned via spoke-map visualizations (Figures 3-8), and improvisation capacity evaluations that distinguish between reactive adaptation and creative problem-solving. This structured analysis reveals where research concentrates, which capabilities remain unrealized, and the specific gaps obstructing advancement toward true collaborative improvisation in construction.

**Level 1: Human-Controlled Execution**

*Current Capabilities and Representative Research*

Level 1 represents complete human dominance in planning and decision-making, with robots functioning as cognitively passive actuators under direct control through teleoperation, preprogramming, or manual supervision (Liang et al., 2021; Saidi et al., 2016). Three control modalities define Level 1: teleoperation through joysticks or AR/VR interfaces for hazardous environments (Chan et al., 2023; Hambuchen et al., 2021), preprogramming for deterministic task reproduction (Khoshnevis et al., 2006), and manual supervision requiring continuous human monitoring (Liu et al., 2017; Schreckenghost et al., 2008). The technological infrastructure privileges mechanical precision over situational autonomy through teleoperation interfaces ranging from simple controllers to sophisticated AR consoles (Xie et al., 2006; Okishiba et al., 2019), deterministic motion control architectures storing and replaying trajectories (Carra et al., 2018), and task-specific end-effectors engineered for particular operations (Bu et al., 2024; Kobayashi et al., 1996).

Representative research spans diverse construction applications. Large-scale additive fabrication exemplifies preprogrammed execution, concrete 3D printing systems like Contour Crafting achieve geometric fidelity through human-supervised deposition paths but struggle with on-site variability (Khoshnevis et al., 2006; Wu et al., 2016; Duballet et al., 2017). Teleoperated construction equipment demonstrates both utility and constraints of remote control, excavator teleoperation enables operators to control heavy machinery from safe locations using emulated control systems with camera and sensor feedback, successfully removing operators from hazardous proximity but concentrating cognitive burden on humans who continuously interpret data and execute corrections without autonomous assistance (Kimura et al., 2006; Lee et al., 2019; Hirabayashi et al., 2006; Kim et al., 2009a). Advanced interfaces incorporating haptic feedback and augmented reality improve performance but preserve human-dominant control (Ootsubo et al., 2015; Okishiba et al., 2019).

Assembly robotics executes preprogrammed manipulation sequences for component installation under human direction, robotic manipulators with specialized grippers handle curtain wall panels, prefabricated modules, and heavy components with geometric precision and load capacity exceeding human capabilities, yet task planning and quality verification remain human responsibilities (Chung et al., 2010; Gambao et al., 2000; Iturralde and Bock, 2013). Infrastructure inspection and maintenance robotics exemplify applications where robots extend human reach into hazardous or inaccessible environments, bridge inspection robots, tunnel inspection systems, and façade cleaning robots operate under teleoperation or follow preprogrammed paths while transmitting sensor data to human operators who interpret findings and direct subsequent actions (Chen et al., 2014; La et al., 2014; Miyake et al., 2006; Pham et al., 2016). Finishing operations including surface treatment and painting employ preprogrammed trajectories or teleoperated commands achieving quality surpassing manual methods yet requiring careful specification and oversight (Pagilla and Yu, 1999; Grassi et al., 2007; Wen and Pagilla, 2018).

Across Levels L1–L6, marker shapes in spoke-maps denote method families: circles (Control/Planning), squares (LfD/imitation), diamonds (RL/IL+RL), triangles (BIM/DT/cloud/knowledge), and hexagons (HRI/interfaces/LLM/trust). Figure 3 and Table 1 show representative Level 1 research primarily extends Physical and, to lesser extent, Cognitive capacity while humans retain Planning and Improvisation dominance. Control- and HRI-focused studies reduce oscillations, improve force regulation, and support safer teleoperation but do not substantially change who decides what to do—cognitive contributions remain limited to better feedback and monitoring with humans providing task intent and improvisation while robots act as precise but largely preprogrammed executors.

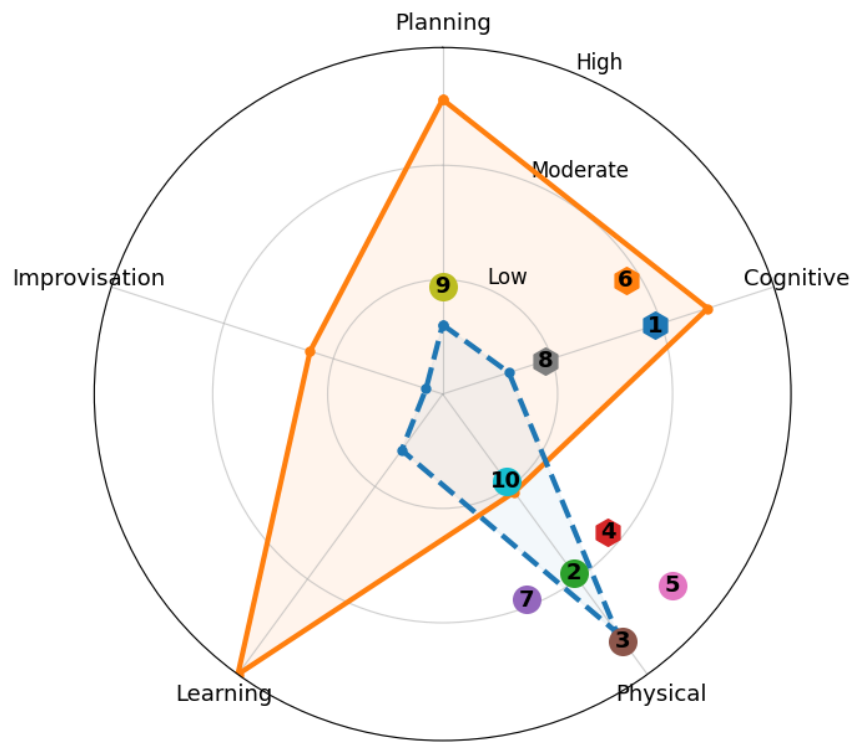

*Fig 3:* Level 1: Human-Controlled Execution – Representative Research Spoke-map

*Improvisation Assessment for Level 1*

By design and architectural necessity, Level 1 systems embody "zero improvisation" from the robotic perspective, fundamentally lacking perceptual, cognitive, and learning capabilities required to perceive novel contingencies, reason about alternative responses, and generate contextually appropriate, previously unplanned behaviors (Musić and Hirche, 2017; Jiang and Arkin, 2015). Specific technical deficits ground this assessment: limited onboard perception and scene interpretation constraining environmental awareness to narrow sensor feedback loops; deterministic motion control lacking contingency reasoning or replanning mechanisms; and negligible knowledge accumulation or learning capabilities preventing generalization across variable site conditions (Robinson et al., 2023; Habibian et al., 2024).

When environments depart from encoded assumptions through unexpected obstacles, material variations, geometric tolerance deviations, or sequencing conflicts, Level 1 systems characteristically halt execution, degrade performance, or await explicit human intervention rather than autonomously recovering (Hambuchen et al., 2021; Böhme and Valenzuela-Astudillo, 2023). Empirical studies document this pattern: concrete printing platforms cannot adjust when material rheology varies, excavators require operator replanning when soil conditions differ, and assembly robots suspend operations when component tolerances exceed programmed acceptance (Wu et al., 2016; Duballet et al., 2017; Zhang et al., 2018).

From a research perspective, Level 1 serves as essential methodological baseline exposing specific capabilities, robust environmental perception, intent inference, policy learning, and shared decision-making that must be introduced to enable robot improvisation and distributed cognition (Jiang and Arkin, 2015; Musić and Hirche, 2017). The "zero improvisation" designation is both descriptive of current technical reality and prescriptive for research trajectories, identifying the locus of technological insufficiency and motivating advancement toward systems that learn from experience, generalize across contexts, and react creatively to unforeseen conditions (Zhang et al., 2025; Habibian et al., 2024).

*Table 1: Level-Human Controlled Execution – Representative Research*

| RepID | Research Focus | Key Citations (Cluster) | Axis (primary skill advanced) | Shape (family) | One-line Key contribution |
|---|---|---|---|---|---|
| 1 | Teleoperation Systems | Goodrich and Schultz (2007); Ootsubo et al. (2015, 2016); Hambuchen et al. (2021); Xie et al. (2006); Chan et al. (2023); Okishiba et al. (2019) | Cognitive | ⬟ | Remote control w/ AR/VR + haptics improves perception & control while keeping human in charge. |
| 2 | Preprogrammed / Fixed-Path Execution | Khoshnevis et al. (2006); Wu et al. (2016); Carra et al. (2018); Robinson et al. (2023); Liu et al. (2017); Schreckenghost et al. (2008) | Physical | ● | CAD-to-trajectory pipelines enable deterministic execution, minimal online adaptation. |
| 3 | 3D Printing / Additive Manufacturing | Khoshnevis et al. (2006); Wu et al. (2016); Duballet et al. (2017); Carneau et al. (2020); Buswell et al. (2022) | Physical | ● | Large-scale deposition (Contour Crafting, concrete AM) extends repeatable placement & reach. |
| 4 | Excavator / Equipment Control | Kim et al. (2009a); Kimura et al. (2006); Lee et al. (2019); Hirabayashi et al. (2006) | Physical | ⬟ | Teleoperated heavy equipment removes operator from hazards; human plans/explains. |
| 5 | Assembly Robotics | Chung et al. (2010); Gambao et al. (2000); Iturralde and Bock (2013); Dharmawan et al. (2017) | Physical | ● | Preprogrammed installers (panels/modules) deliver precision & load capacity under supervision. |
| 6 | Infrastructure Inspection & Maintenance | Chen et al. (2014); La et al. (2014); Pham et al. (2016); Miyake et al. (2006) | Cognitive | ⬟ | Mobile sensing platforms collect data in hard-to-reach places; humans interpret & decide. |
| 7 | Finishing Operations | Pagilla and Yu (1999); Grassi et al. (2007); Kobayashi et al. (1996); Wen and Pagilla (2018); Gracia et al. (2019) | Physical | ● | Preplanned finishing/painting/troweling improves consistency; reasoning remains human. |
| 8 | Safety & HRI | Lasota et al. (2017); Böhme and Valenzuela-Astudillo (2023); Liang et al. (2021) | Cognitive (UI, safety logic) | ⬟ | Safe-HRI/MR tooling raises situational awareness but preserves human-dominant control. |
| 9 | Control Architecture & Frameworks | Musić and Hirche (2017); Jiang and Arkin (2015); Habibian et al. (2024) | Planning | ● | Mixed-initiative/autonomy-level frameworks structure supervision without autonomous planning. |
| 10 | End-Effectors & Specialized Tools | Bu et al. (2024); Kobayashi et al. (1996) | Physical | ● | Task-specific tooling (weld/trowel) increases execution capability, no situational reasoning. |

**Level 2: Adaptive Manipulation Systems**

*Current Capabilities and Representative Research*

Level 2 represents the transitional stage where robots incorporate sensor-driven adaptation and closed-loop feedback mechanisms enabling localized autonomous adjustments beyond purely human-controlled execution (Liang et al., 2021). Systems integrate multimodal sensing such as machine vision, depth sensing, LiDAR, force/torque sensors, tactile feedback with real-time control algorithms facilitating micro-level corrections to motion and interaction parameters (Haddadin and Shahriari, 2024; Rodríguez-Martínez et al., 2025; Shao et al., 2019; Kaczmarek, 2019). This fusion enables automatic correction of minor positioning errors, contact force regulation, and operational stability maintenance amidst environmental variations without continuous human input (Abu-Dakka and Saveriano, 2020; Keemink et al., 2018). The technical hallmark is closed-loop control architectures where sensing, planning, and actuation cycles operate autonomously within predefined operational envelopes (Liang et al., 2021). Variable impedance control paradigms enable online modulation of robot dynamics during contact tasks by adjusting stiffness and damping in response to environmental cues (Haddadin and Shahriari, 2024; Caldarelli et al., 2022). Visual servoing systems use image-based feedback for real-time pose correction (Moughlbay et al., 2013), while force-aware controllers maintain desired contact conditions during manipulation (Li et al., 2023; Cao et al., 2020). Motion planning algorithms reduce oscillations and improve trajectory stability under environmental variations (Wang et al., 2023).

Operationally, Level 2 systems perform autonomous micro-adjustments, fine positional refinements, trajectory re-timings, and local force modulations while relying on humans for high-level planning, decision-making, exception management, and task redefinition (Lundeen et al., 2017, 2019; Feng et al., 2015). The human role shifts from continuous teleoperation toward supervisory monitoring and intermittent intervention when tasks exit stable operational envelopes (Pervez et al., 2019; Tung et al., 2020). This pattern reduces low-level motor workload but increases demands for situational awareness and rapid intervention when bounded autonomy fails (You et al., 2018; Rozo et al., 2016). Construction-relevant implementations combine industrial manipulators with perception suites and force sensing for on-site fabrication and assembly. Learning-from-demonstration approaches enable robots to acquire geometrically adaptive construction subtasks, reproducing primitives while adapting to geometric variations (Liang et al., 2019, 2020). Feedback-enabled digital fabrication systems employ vision and depth sensing to monitor and correct material deposition (Lundeen et al., 2019), while adaptive platforms use stereo vision and force feedback for quasi-repetitive tasks accommodating component tolerances and surface variability (Wang et al., 2022; Rozo et al., 2013).

Representative research occupies several complementary methodological families. Vision-guided manipulation studies demonstrate how image-based feedback closes control loops for pose correction and alignment essential for construction assembly (Feng et al., 2015; Lundeen et al., 2017). Scene understanding systems show how robots autonomously adjust work plans based on perceived environmental conditions, though within bounded parameters established during programming (Lundeen et al., 2017, 2019). Force-based learning from demonstration frameworks document progress in contact-rich tasks where robots learn insertion and compliant assembly by regulating interaction forces based on sensory feedback (Rozo et al., 2013, 2016; Wang et al., 2022; Zeng et al., 2019). These approaches combine trajectory learning with force and impedance policies, allowing local error minimization and compliant adaptation responding to minor misalignments (Liang et al., 2020). Haptic exploration schemes and local recovery models exemplify how Level 2 robots detect and react to deviations, probing contact surfaces to select compliant strategies or executing learned recovery maneuvers when small failures are detected (Eiband et al., 2019; Duchetto et al., 2018), though these remain effective only when failure modes fall within prior experience scope and sensor observability (Lundeen et al., 2019).

Shared teleoperation and multimodal teaching research combining visual demonstrations, kinesthetic guidance, and haptic feedback illustrates Level 2's transitional character, where these modalities enable robots to acquire primitives later adapted with closed-loop controllers, reducing operator burden while leaving strategic coordination to humans (Tung et al., 2020; Pervez et al., 2019). Construction-specific implementations demonstrate teaching robots' quasi-repetitive tasks through human demonstration, where learned motion primitives execute with sensor-mediated corrections for geometric variations (Liang et al., 2020).

Perception integration with control demonstrates vision system maturity feeding adaptive control loops. Convolutional neural networks for real-time construction environment detection supply situational information such as object poses, obstacle locations, human presence required for localized adaptation and safety-aware intervention (Hayles et al., 2022; You et al., 2018). Stereo vision and depth sensing enable pose estimation and local surface characterization supporting adaptive manipulation, though performance remains bounded by sensor fidelity and environmental conditions (Shao et al., 2019; Kaczmarek, 2019). Advances in stereo and LiDAR fusion alongside visual-inertial simultaneous localization and mapping provide the metric fidelity and pose information needed for effective control in semi-structured environments (Shao et al., 2019; Sun et al., 2024). Frameworks for multimodal encoding and sensor fusion demonstrate how heterogeneous inputs combining vision with tactile or force data are used to parameterize and adapt manipulation skills for contact-rich or geometrically variable tasks (Zeng et al., 2019; Wang et al., 2022).

Human-robot safety, interface design, and immersive training research addresses the psychosocial and ergonomic conditions under which supervisory control rather than direct control becomes viable. You et al. (2018) explore immersive VR to enhance perceived safety, documenting how training and interface design influence operator intervention ability. Separated workspace design promotes worker trust and reduces fear (You et al., 2018; Liang et al., 2021). Haptic-guided control enables online stiffness inference and contact adaptation (Caldarelli et al., 2022; Eiband et al., 2019). Research on haptic-guided and perturbation-based control methods demonstrates how online stiffness inference and contact adaptation can maintain stability during slight environmental variations (Caldarelli et al., 2022; Eiband et al., 2019). Variable impedance and compliant control research establish unified frameworks for force-impedance regulation. Contemporary advances synthesize these approaches into unified control architectures enabling seamless transitions between different control modes (Haddadin and Shahriari, 2024). Admittance control frameworks for physical human-robot interaction demonstrate how robots can safely respond to human forces during collaborative tasks (Keemink et al., 2018). Applications in robot surface finishing, grinding, and polishing show force tracking and adaptive impedance control strategies enabling consistent quality across varying workpiece geometries (Li et al., 2018; Yang et al., 2017).

Motion planning and low-level control research addresses trajectory generation and execution stability. Path-planning methods significantly reduce local oscillations in manipulator movements, critical for construction tasks requiring smooth, predictable motion (Wang et al., 2023). Integration of motion planning with learned manipulation policies demonstrates how geometric constraints and workspace boundaries can be encoded to ensure safe, efficient task execution (Wang et al., 2022; Liang et al., 2021). Research on motion encoding and trajectory representation shows how teleoperation traces can be captured and reproduced with variations accommodating different workspace configurations (Lundeen et al., 2019). Constraint modeling and foundational work explicitly distinguishing mechanical adaptation from creative improvisation provide critical conceptual framing for Level 2 capabilities. Fitzgerald et al. (2021) explicitly distinguish mechanically feasible substitutions achievable through sensor-driven control from cognitive creativity underlying human improvisation, highlighting the gap between Level 2's error-minimizing adaptation and reasoning capacities for unconstrained improvisation.

At Level 2, Figure 4 and Table 2 indicate that representative work concentrates on enhancing Physical execution and Cognitive feedback, with early gains in Learning but only modest change in Improvisation. Circles for control/force work and hexagons for perception/HRI dominate the Physical and Cognitive axes, while a smaller group of LfD squares begins to populate the Learning axis. Motion planning, compliant/impedance control, and perception-state estimation give robots more robust, adaptive contact-rich behavior, but planning authority and creative problem solving remain largely with the human, who supervises, intervenes, and re-directs the robot when conditions fall outside the learned envelope.

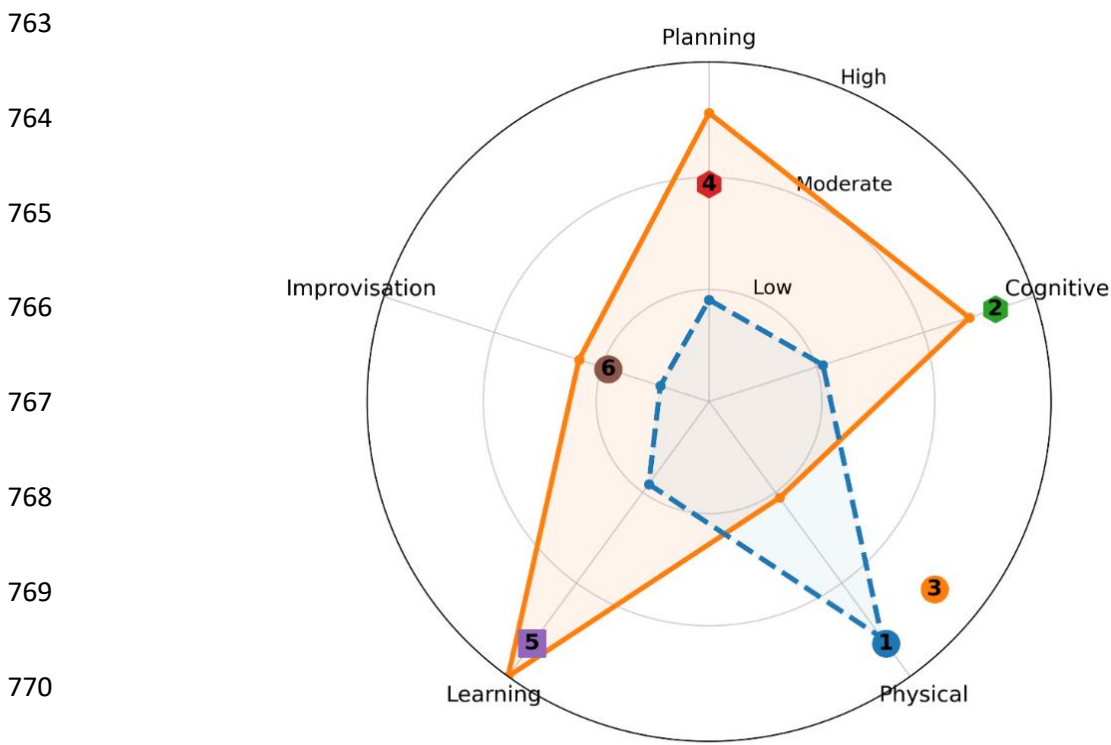

***Fig 4:*** *Level 2: Adaptive Manipulation Systems – Representative Research Spoke-map*

*Table 3: Level 2 – Adaptive Manipulation Systems: Representative Research Table*

| RepID | Research focus | Key citations | Axis (primary skill advanced) | Shape (Family) | One-line Key contribution |
|---|---|---|---|---|---|
| 1 | Motion Planning and Low-Level Control | Wang et al. (2023); Wang et al. (2022); Liang et al. (2021); Lundeen et al. (2019) | **Physical** (sec: Cognitive) | ● | Motion-planning and low-level control reduce oscillations and enable geometrically adaptive, feedback-driven execution. |
| 2 | Perception and Real-Time State Estimation | Hayles et al. (2022); You et al. (2018); Shao et al. (2019); Sun et al. (2024); Rodríguez-Martínez et al. (2025); Kaczmarek (2019); Moughlbay et al. (2013) | **Cognitive** | ⬟ | CNNs, SLAM, and multi-sensor fusion provide real-time scene understanding for closed-loop adaptive manipulation. |
| 3 | Variable Impedance and Compliant Control | Haddadin & Shahriari (2024); Li et al. (2018); Keemink et al. (2018); Caldarelli et al. (2022) | **Physical** | ● | Unified force–impedance and compliant control regulate contact forces during contact-rich, adaptive manipulation. |
| 4 | Human–Robot Safety and Interface Design | You et al. (2018); Caldarelli et al. (2022); Eiband et al. (2019); Liang et al. (2021) | **Planning** (supervisory) | ⬟ | Safety logic and immersive interfaces support supervisory control and timely human intervention during adaptation. |
| 5 | Learning from Demonstration | Rozo et al. (2013); Wang et al. (2022) | **Learning** (sec: Physical) | ■ | LfD frameworks teach adaptive assembly/insertion primitives that generalize across similar contact-rich tasks. |
| 6 | Constraint Modeling | Fitzgerald et al. (2021); Eiband et al. (2019); Duchetto et al. (2018) | **Improvisation** | ● | Constraint models clarify feasible substitutions and highlight the gap between feedback adaptation and true improvisation. |

*Improvisation Assessment for Level 2*

Assessing Level 2 improvisational capacity requires distinguishing between mechanical, feedback-driven adaptation—reactive, bounded, and parameterized and cognitive improvisation characterized by anticipation, generative reasoning, and context-sensitive problem-solving (Liang et al., 2021). Level 2 systems predominantly manifest the former: adapting only through reactions to sensed perturbations within coded thresholds and pre-specified recovery strategies (Fitzgerald et al., 2021). Control algorithms autonomously alter trajectories or maintain contact forces within bands, but responses are explicitly constrained by algorithmic design choices and sensor models defining permissible corrective actions (Wang et al., 2023; Haddadin and Shahriari, 2024).

Practical limitations curtail improvised behavior. First, operational envelopes are explicitly encoded through parameter ranges, safe-stop logics, and recovery policies maintained conservatively for safety (Liang et al., 2021). Second, sensor and model uncertainties including noise, occlusion, and domain shifts produce failure modes requiring human task redefinition or replanning (Wang et al., 2023; Hayles et al., 2022; You et al., 2018). Third, software architectures separate feedback-centric control from representational learning needed for contextual generalization, lacking planning and analogical reasoning underpinning human improvisation (Hayles et al., 2022; You et al., 2018; Fitzgerald et al., 2021).

Robots recover from anticipated deviations like slight misalignments via sensor-guided policy selection, but recoveries remain limited to scenarios in training data or modeled primitives (Duchetto et al., 2018; Wang et al., 2022; Rozo et al., 2013). When perturbations exceed sensors' observability or require subtask goal redefinition, human operators intervene to diagnose, replan, or improvise (Liang et al., 2021). Human workers employ flexible reasoning, contextual sense-making, analogical tool substitution, and social coordination, relying on rich internal models and cross-modal reasoning capabilities exceeding Level 2's local feedback correction (Hayles et al., 2022; You et al., 2018; Fitzgerald et al., 2021; Mohan et al., 2019).

Critically, Level 2 establishes sensory, control, and algorithmic prerequisites permitting transition toward higher taxonomy levels. Closed-loop perception modules, force sensing, and trajectory control methods deployed at Level 2 create infrastructural substrate upon which learning from demonstration, policy generalization, semantic task representations, and shared reasoning can be layered (Liang et al., 2021; Osa et al., 2018). Work on robotic learning of assembly tasks via human demonstration builds on sensors and local controllers similar to

Level 2, indicating natural progression: improving perception and control robustness reduces demonstration replay brittleness and supports more effective generalization in subsequent levels (Liang et al., 2020; Wang et al., 2022). Level 2 thus represents the crucial transitional milestone between direct human control and the emergence of machine-mediated learning, consultation, and ultimately joint improvisation characterizing higher collaboration levels. Thus, Level-2 perception/control infrastructures are the substrate upon which LfD, policy generalization, and shared reasoning develop at higher levels.

**Level 3: Learning from Demonstration (LfD) Systems**

Level 3 represents paradigm shift from sensor-driven adaptive manipulation toward learning-based policy acquisition where robots internalize task models from human demonstrations rather than executing hand-coded programs or reactive sensor routines (Osa et al., 2018; Liang et al., 2020). At this level, robots learn mappings from observed demonstrations to control actions through kinesthetic guidance, teleoperation traces, or visual observation, inferring policies that can be reproduced and interpolated across similar conditions (Tai et al., 2016). LfD systems accept multiple demonstration modalities: kinesthetic teaching through physical guidance, teleoperation traces from expert remote control, and observational demonstrations captured as video sequences (Zeng et al., 2019; Torabi, 2019). Learning from observation (LfO) proves increasingly feasible through representation learning, enabling LfD when action labels are unavailable or costly to record (Kim et al., 2022; Sermanet et al., 2018). Multimodal demonstrations encoding vision, proprioception, and haptics form richer skill representations improving policy robustness in partially observed scenes (Zeng et al., 2019; Si et al., 2022).

Foundational algorithmic work spanning behavioral cloning, inverse-reinforcement learning, adversarial imitation, on-policy correction techniques like DAgger, and representation learning from video provides methodological substrate for Level 3 (Osa et al., 2018; Kostrikov et al., 2018; Ross and Bagnell, 2014). C-LEARN demonstrated learning geometric constraints from demonstrations to support multi-step manipulation and shared autonomy, showing how constraint extraction structures policy generalization for multi-stage tasks (Pérez-D'Arpino and Shah, 2017). Work on corrective demonstration and multi-resolution representations formalized how hierarchical demonstrations enable efficient policy refinement and error recovery (Cai et al., 2020; Meriçli et al., 2012). Time-contrastive and self-supervised representation methods enable robots to extract task-relevant embeddings from demonstrations recorded in various views, facilitating subsequent policy learning (Sermanet et al., 2018).

Human pedagogical factors matter, first-person demonstrations improve transfer by aligning perceptual frames (Fiorella et al., 2017; Ransom et al., 2022). Interactive on-policy methods reduce compounding errors by enabling supervisors to provide corrective labels for states visited by robot policies, improving robustness for sequential tasks (Zhang and Cho, 2017; Lee et al., 2021; Balakrishna et al., 2019). Hierarchical imitation formulations learn subtask policies from unsegmented demonstrations, facilitating decomposition of long-horizon sequences into reusable primitives desirable for construction workflows composed of repeated subtasks (Sharma et al., 2018; Gupta et al., 2019).

Construction-relevant applications demonstrate teaching robots quasi-repetitive tasks through human demonstration where learned motion primitives execute with sensor-mediated corrections for geometric variations (Liang et al., 2019, 2020). Integrated pipelining of teleoperation or kinesthetic capture with variable impedance control shows effective human skill transfer for contact-sensitive tasks, preserving critical force-sensitive aspects during autonomous replay (Jiang et al., 2021; Abu-Dakka et al., 2014; Si et al., 2022). Studies combining imitation with reinforcement learning demonstrate using demonstration priors to initialize policies while enabling experiential refinement to enhance precision when execution deviations require corrective adaptation (Wang et al., 2020; Kormushev et al., 2010). Work robustifying LfD pipelines through Bayesian disturbance injection or stationary-distribution correction addresses noise, partial observability, and model misspecification endemic to construction settings (Oh et al., 2022; Kim et al., 2022). Federated imitation learning frameworks show how heterogeneous demonstration data from distributed workers can be aggregated to produce more generalizable policies valuable for construction where demonstrations occur across different sites and hardware configurations (Beltran-Hernandez et al., 2020).

Figure 5 and Table 3 show a clear shift toward the Learning axis, with strong contributions from foundational LfD algorithms, learning-from-observation, and meta-imitation methods. Representative work (RepIDs in Table 3) enables robots to acquire richer, more transferable manipulation skills from demonstrations, including some hierarchical and IL+RL hybrids that begin to touch Planning and Improvisation. Nevertheless, human workers still frame tasks, choose which skills to invoke, and arbitrate unusual situations, so the robot's "creativity" is bounded by the demonstration space and the scenarios encoded in the training distribution.

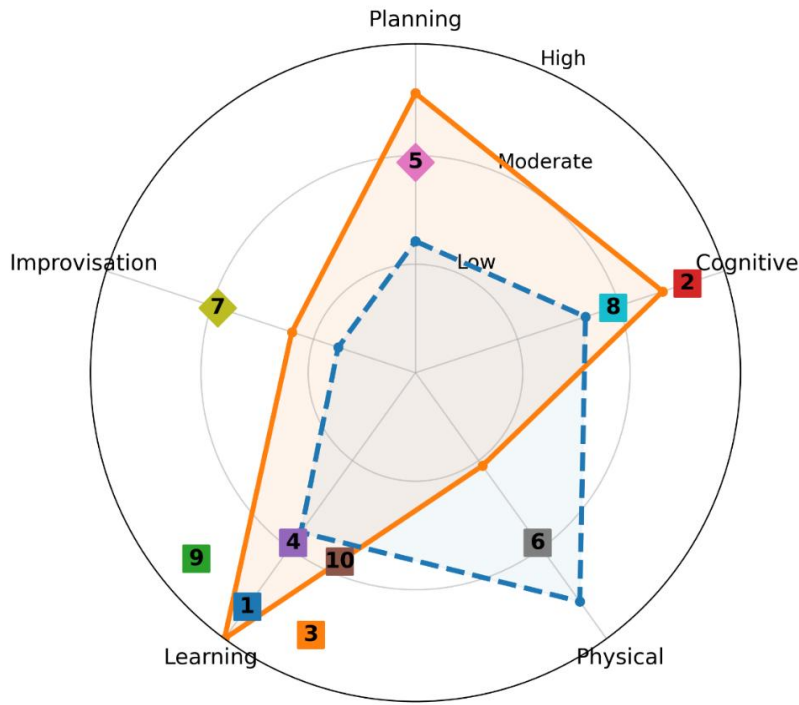

*Fig 5:* Level 3: Learning from Demonstration Systems – Representative Research Spoke-map

*Improvisation Assessment for level 3*

LfD at Level 3 produces circumscribed improvisation: robots interpolate or combine known behaviors for effective executions under modest perturbations but remain constrained by demonstration distributions and learned representation expressivity (Kim et al., 2022; Viano et al., 2022). Level 3 improvisation is retrospective and example-driven rather than generative—robots synthesize variations near past exemplar manifolds but do not invent novel strategies absent from training data (Osa et al., 2018). Imitation-based learners acquire mappings from demonstrations but rarely possess task semantics models or higher-order reasoning mechanisms supporting creative re-planning in truly novel problem contexts (Breazeal and Scassellati, 2000; Hajimirsadeghi et al., 2012).

A primary limitation is that LfD derives behavior from past exemplars instead of actively synthesizing prospective, model-based alternatives needed for creative improvisation (Jiang et al., 2021; Devin et al., 2017). While hybridization with reinforcement learning moves systems toward prospective adaptation, these methods still require significant online data or carefully designed simulators to achieve novelty generation beyond interpolation (Kormushev et al., 2010; Anand et al., 2022). Construction-specific studies reveal demonstration quality and coverage critically constrain performance insufficient demonstrations yield brittle behavior, and when perturbations require subtask goal redefinition such as altered assembly sequences,

Level 3 systems characteristically fail and require human intervention (Liang et al., 2020; Pérez-D'Arpino and Shah, 2017).

However, Level 3 is significant because it establishes first substantive cognitive overlap between human and robot collaborators: robots internalize task patterns and reason in experience-based policy spaces while humans adapt teaching strategies to shape those policy spaces, reciprocity central to higher HRC levels (Hu et al., 2020; Jiang et al., 2024). Level 3 therefore replaces programmed or reactive execution with empirical, demonstration-centered task competence acquisition, enabling robots to participate more actively in shared problem-solving while remaining bounded by demonstration data (Osa et al., 2018). The algorithmic advances, which includes IfO, hierarchical imitation, on-policy correction, IL+RL hybridization which provide methodological foundations for Level 4 systems where human-in-the-loop BIM or digital twin contexts afford richer semantic grounding (Gupta et al., 2019; Beltran-Hernandez et al., 2020).

*Table 3: Level 3 – Learning from Demonstration: Representative Research Summary*

| RepID | Research focus | Key citations (Cluster) | Axis (primary skill advanced) | Shape (Family) | One-line key contribution |
|---|---|---|---|---|---|
| 1 | Foundational LfD algorithms | Osa et al. (2018); Kostrikov et al. (2018); Ross & Bagnell (2014); Sun et al. (2022); Tai et al. (2016) | **Learning** (sec: Cognitive) | ■ | Provides the core algorithmic toolkit for imitation learning, from behavioral cloning to IRL/adversarial and interactive IL. |
| 2 | Constraint learning & multi-step manipulation | Pérez-D'Arpino & Shah (2017); Cai et al. (2020); Meriçli et al. (2012) | **Cognitive** (task structure) | ■ | Learns geometric/task constraints from demonstrations, enabling multi-step, structured manipulation and shared autonomy. |
| 3 | Learning from observation (IfO) & visual learning | Torabi (2019); Kim et al. (2022); Sermanet et al. (2018); Fiorella et al. (2017); Ransom et al. (2022) | **Learning** | ■ | Shows how robots can learn from video/observation alone via representation learning and time-contrastive embeddings. |
| 4 | On-policy & interactive imitation | Lee et al. (2021); Zhang & Cho (2017); Balakrishna et al. (2019) | **Learning** (robustness) | ■ | Reduces compounding errors through on-policy correction and interactive labeling along robot-visited trajectories. |
| 5 | Hierarchical & long-horizon imitation | Sharma et al. (2018); Gupta et al. (2019) | **Planning** (hierarchy) | ◆ | Learns hierarchical policies and relay structures, decomposing long-horizon tasks into reusable subtask policies. |
| 6 | Construction & contact-rich applications | Liang et al. (2019, 2020); Zeng et al. (2019); Si et al. (2022); Abu-Dakka et al. (2014); Jiang et al. (2021) | **Physical** (sec: Learning) | ■ | Demonstrates LfD for quasi-repetitive, contact-rich subtasks (e.g., assembly, peg-in-hole) with sensor-mediated corrections. |
| 7 | Hybrid IL + RL approaches | Wang et al. (2020); Kormushev et al. (2010); Devin et al. (2017) | **Improvisation** (via RL) | ◆ | Combines demonstration priors with RL to refine skills and modestly expand behavior beyond pure interpolation. |
| 8 | Robustness & model misspecification | Oh et al. (2022); Kim et al. (2022); Viano et al. (2022) | **Cognitive** (robust policy) | ■ | Addresses noise, partial observability, and model mismatch to make LfD policies more reliable under real-world disturbances. |
| 9 | Meta-learning & few-shot imitation | Hu et al. (2020); Jiang et al. (2024) | **Learning** | ■ | Uses meta-learning to enable rapid adaptation and one-shot imitation from very few demonstrations. |
| 10 | Federated & distributed imitation | Beltran-Hernandez et al. (2020) | **Learning** | ■ | Explores aggregating heterogeneous demonstrations across workers/sites to train more generalizable policies. |

**Level 4: Human-in-Loop BIM-driven Systems**

*Current Capabilities and Representative Research*

Level 4 represents the transition from task-level experiential learning toward context-aware, model-driven autonomy, where robots operate within semantically rich digital construction models such as Building Information Modeling (BIM) and Digital Twins under human supervisory control (Ramasubramanian et al., 2022; Vieira et al., 2023). BIM and Digital Twins form the informational backbone shared between digital planners, human supervisors, and on-site automated agents, enabling closed-loop execution informed by design data, scheduling constraints, and sensor observations (Dai et al., 2024; Coupry et al., 2021). Digital Twin architectures integrate geometric, semantic (BIM/IFC), and temporal data streams with operational telemetry, creating an information substrate on which robots resolve spatial relationships, reference design intent, and monitor process variables (Dai et al., 2024; Vieira et al., 2023). BIM models provide canonical representations of design geometry and metadata, while field sensing scan data, Real-Time Location Systems (RTLS), Internet of Things (IoT) telemetry which supplies as-built state that can be matched to the model (Chancharoen et al., 2022). Deviations are detected algorithmically and routed into human-supervised decision loops for correction or re-planning (Kavaliauskas et al., 2022).

Technical enablers include high-fidelity geometric acquisition through terrestrial/mobile LiDAR, photogrammetry, and Unmanned Aerial Vehicle (UAV) scanning with automated point-cloud-to-BIM conversion; sensor fusion architectures and SLAM pipelines for mobile robotic localization within model coordinates; and IFC-compliant semantic frameworks exposing object properties, task sequences, and schedule information to reasoning agents (Macher et al., 2017; Zeng et al., 2023; Jung et al., 2015). Progressive laser scanning and SLAM-based mapping permit construction environments to be captured iteratively and integrated into BIM, supporting on-site decision making and automated verification against design intent (Matellon et al., 2023; Frías et al., 2019). Mixed-reality and immersive interfaces extend human supervision by visualizing proposed robot actions in situ and enabling lightweight authoring of corrections or constraints immediately consumable by robot controllers and shared models (Coupry et al., 2021; Jahnke et al., 2023). Within this framework, human operators occupy cognitive and normative roles: they validate robot-generated proposals, adjudicate safety and ethical trade-offs, resolve semantic conflicts between model and sensor evidence, and supply creative problem solving when conditions fall outside encoded model scope (Hasan and Sacks, 2021; Ramasubramanian et al., 2022).

Research demonstrating Level 4 configurations embeds robots within BIM/Digital Twin environments, showing bidirectional human-robot-model interaction and model-based correction. Immersive industrialized construction environments illustrate how virtual environments fed by BIM/Digital Twin data enable training and real-time coordination between remote experts, human crews, and automated assets (Podder et al., 2022; Javerliat et al., 2023). Digital Twin architectures articulate how twins serve as media for integrating design intent, sensor streams, and control logic underpinning human-supervised autonomy (Song et al., 2022; Hause, 2019).

Scan-to-BIM workflows supply as-built geometries and semantic element information situating robot actions within modeled contexts, supporting comparisons between as-designed and as-built states for deviation detection and corrective planning (Macher et al., 2017; Kavaliauskas et al., 2022). Studies demonstrate that 3D laser scanning meaningfully improves as-is model generation with progressive scanning strategies maintaining up to date as-built representations during active construction (Frías et al., 2019; Sadeghineko et al., 2024; Abreu et al., 2023). BIM-derived geometric and semantic maps serve as high-level priors for mobile robot localization, enabling robots to register local SLAM-based maps against BIM coordinate frames and ground task execution in design semantics rather than purely metric maps (Jung et al., 2015, 2016; Yin et al., 2022). These capabilities are essential for robots to interpret as-designed versus as-built conditions and reason about element-level tasks (Jung et al., 2015).

Experimental studies demonstrate model-augmented control in construction settings. Mixed-reality assisted assembly using 3D virtual models and Convolutional neural networks (CNN) based perception shows robots can be directed by model-anchored perception pipelines while human supervisors leverage mixed-reality displays to monitor and correct execution (Židek et al., 2021; Chheang et al., 2024). Digital-twin-driven welding workstations predict process quality and feed corrective control commands back to physical robots, while collaborative painting robots use twins to simulate motion and diagnose deviations, proposing parameter updates vetted by human operators (Zhang et al., 2023; Chancharoen et al., 2022). Steel-structure digital twins and IFC-based modeling show pathways for encoding assembly sequences, tolerances, and logistics constraints in machine-readable form (Liu and Lin, 2024).

Studies addressing automatic or semi-automatic detection of deviations between as-designed BIM and as-built point clouds produce algorithms flagging non-conformances and proposing corrective measures for human review (Jafari et al., 2020; Kavaliauskas et al., 2022). Research on automated quality inspection shows that terrestrial laser scanning combined with BIM enables rapid identification of geometric errors and can feed back to on-site actors or robotic manipulators for corrective interventions (Wang et al., 2015; Bariczová et al., 2021). Mixed-reality and visualization modalities emphasize intuitive interfaces enabling human supervisors to interpret robot-generated model comparisons and validate or adjust autonomous proposals (Williams et al., 2018; Lalik and Flaga, 2021). Prototypical systems show mixed reality can bridge gaps between digital twins and physical sites by overlaying BIM elements, detected deviations, and proposed corrections in situ, shortening cognitive loops for human validation (Williams et al., 2018).

Figure 6 and Table 4 highlight a strong emphasis on Planning and Cognitive capacity through BIM/DT-centered methods. Digital twin architectures, scan-to-BIM pipelines, deviation detection, and BIM-based localization create a shared, up-to-date model of the site that both humans and robots can consult, while MR interfaces and immersive DT environments support richer human–robot coordination. Robots become more proactive in proposing actions that are consistent with the twin, but humans still validate changes, prioritize goals, and decide which deviations warrant intervention, so model-mediated collaboration reinforces rather than replaces human planning.

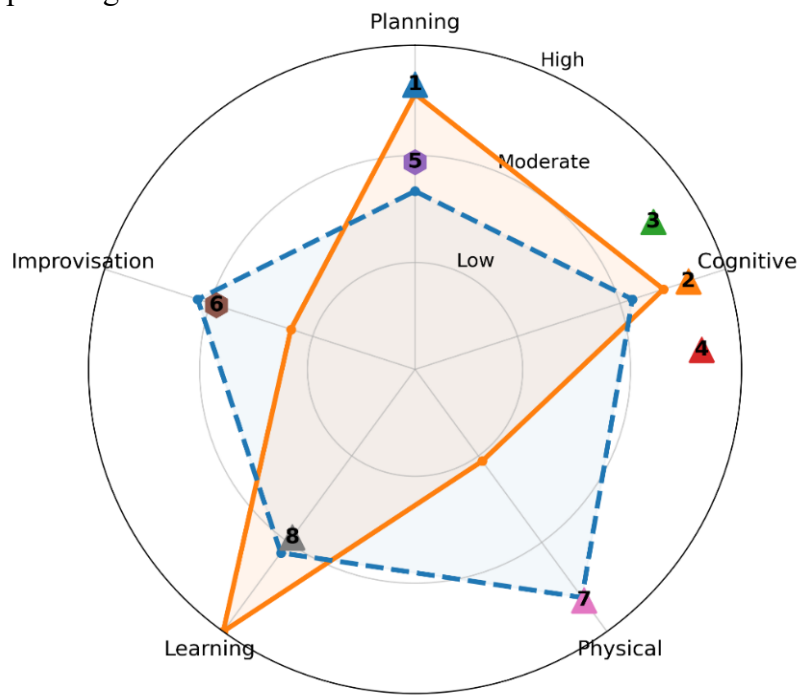

***Fig 6:*** *Level 4: Human-in-Loop BIM-driven Systems – Representative Research Spoke-map*

*Table 4: Level 4: Human-in-Loop BIM-driven Systems: Representative Research Summary*

| RepID | Research focus | Key citations (cluster) | Axis (primary skill advanced) | Shape (Family) | One-line key contribution |
|---|---|---|---|---|---|
| 1 | Digital Twin Architectures and Foundations | Hause (2019); Song et al. (2022); Vieira et al. (2023); Dai et al. (2024); Ramasubramanian et al. (2022) | **Planning** | ▲ | Defines DT architectures that integrate design, process, and telemetry, enabling model-driven planning under human supervision. |
| 2 | Scan-to-BIM and As-Built Acquisition | Macher et al. (2017); Frías et al. (2019); Zeng et al. (2023); Abreu et al. (2023); Sadeghineko et al. (2024); Matellon et al. (2023) | **Cognitive** | ▲ | Develops point-cloud-to-BIM and progressive scanning workflows so robots and twins have accurate, up to date as-built context. |
| 3 | BIM-Based Robot Localization and SLAM | Jung et al. (2015, 2016); Yin et al. (2022) | **Cognitive** | ▲ | Uses BIM-derived maps as priors for SLAM and semantic localization, grounding robot pose and navigation in design coordinates. |
| 4 | Automated Deviation Detection & Quality Inspection | Wang et al. (2015); Jafari et al. (2020); Bariczová et al. (2021); Kavaliauskas et al. (2022) | **Cognitive** | ▲ | Compares scans to BIM to flag mismatches and quality issues, feeding human-supervised corrective planning and robot interventions. |
| 5 | Mixed Reality and Human Interfaces | Williams et al. (2018); Lalik & Flaga (2021); Coupry et al. (2021); Jahnke et al. (2023) | **Planning** (supervisory) | ⬟ | MR interfaces visualize BIM/DT data and robot proposals in situ, supporting intuitive human supervision and plan adjustment. |
| 6 | Immersive Digital Twins & Collaborative Environments | Podder et al. (2022); Javerliat et al. (2023); Chheang et al. (2024) | **Improvisation** (team-level) | ⬟ | Immersive digital twins enable remote experts and crews to jointly explore alternatives and coordinate corrective actions with robots. |
| 7 | Digital Twin–Driven Robotic Applications | Chancharoen et al. (2022); Zhang et al. (2023); Židek et al. (2021); Liu & Lin (2024) | **Physical** | ▲ | Uses digital twins to simulate, optimize, and correct welding, painting, and assembly robots before executing human-approved adjustments on site. |
| 8 | BIM/DT Integration Architectures | Coronado et al. (2024); Foo et al. (2023); Hasan & Sacks (2021) | **Learning** | ▲ | Proposes integration paths that let organizations accumulate and reuse BIM/DT knowledge across projects, supporting longer-term system learning. |

*Improvisation Assessment for Level 4*

Level 4 introduces moderate but principled improvisational capability grounded in model-informed reasoning and human-guided adaptation. Robots detect discrepancies between as-designed intent and as-found conditions and generate corrective proposals, replanning motion paths, suggesting alternative sequences, or proposing parameter adjustments within pre-specified safety and task constraints, evaluated by human supervisors before enactment (Zhang et al., 2023; Chancharoen et al., 2022). This capacity for model-anchored corrective generation constitutes qualitative enhancement over Level 3's experiential imitation: robots reason about semantic task structure, spatial relationships, and permissible deviations, formulating adaptation hypotheses interpretable to human collaborators (Ramasubramanian et al., 2022; Vieira et al., 2023). Where Level 3 produces behavior generalizing from exemplars, Level 4 affords contextual adaptation: robots' reason about geometry and task dependencies, re-sequence installation steps when preceding elements are out of tolerance, propose alternative task allocations or motion plans, and present model-aligned options for human ratification (Kavaliauskas et al., 2022; Jafari et al., 2020).

Nevertheless, improvisation remains bounded and structured. Robot proposals are constrained by model fidelity and semantic richness of BIM/IFC data, accuracy and timeliness of as-built acquisition, and robustness of perception and registration algorithms (Hajian and Becerik-Gerber, 2010; Zeng et al., 2023). Operational errors which includes registration inaccuracies, segmentation difficulties, semantic misclassification which reduce reliability of automated deviation detection and limit scope of safe autonomous correction (Abreu et al., 2023; Park et al., 2024). Institutional factors including intermittent sensor coverage and organizational change management practices also limit robot improvisation in live construction settings (Uotila et al., 2021; Hasan and Sacks, 2021).

Human roles shift from low-level teleoperator to cognitive collaborator and arbiter: validating robot-proposed adaptations, resolving ambiguities when digital and real conditions diverge, and supplying domain knowledge not encoded within BIM schemas such as tacit workmanship practices, contractual risk judgments, contextual safety tradeoffs (Williams et al., 2018; Mejlænder-Larsen, 2019; Coupry et al., 2021).

Level 4 represents qualitative change from experiential adaptation to contextual adaptation through model-informed reasoning constrained by encoded design intent and measured reality (Ramasubramanian et al., 2022). This hybrid improvisation, which is bounded, explainable, and human-validated, advances improvisational capacity by institutionalizing shared situational awareness: robots participate in structured improvisational cycles identifying anomalies, suggesting constrained alternatives, and executing human-approved corrections, while humans remain principal sources of creative problem-solving (Kavaliauskas et al., 2022; Dai et al., 2024).

**Level 5: Cloud-Based Knowledge Systems (Current Research Frontier)**

*Current Capabilities and Representative Research*

Level 5 marks a conceptual shift from project-bound, model-driven autonomy toward cloud-connected, knowledge-sharing robotic systems that treat knowledge as a distributed resource (Kehoe et al., 2015; Firoozi et al., 2024). Where Level 4 systems rely on a project's BIM instance and local human validation, Level 5 systems augment on-site models with remotely hosted repositories, foundation-scale AI services, and federated knowledge stores continuously updated and shared across teams and projects (Afsari et al., 2016; Wan et al., 2016). Autonomy is no longer bounded by a single project model but by access to a growing, externally curated corpus of experience and reasoning primitives (Firoozi et al., 2024).

Architecturally, Level 5 robots' access three interlocking knowledge strata: collective data repositories (cloud BIM, sensor archives, annotated case histories), cloud-hosted AI models offering cross-domain reasoning, and distributed knowledge networks enabling peer-to-peer exchange of policies and lessons learned (You et al., 2023; Firoozi et al., 2024). These strata enable two complementary modes: horizontal consultation which is peer-to-peer exchange among robotic agents sharing learned control policies, constraints, or telemetry summaries across fleets and vertical consultation that is retrieving authoritative human, institutional, or standards knowledge through cloud interfaces, such as querying standards databases or expert-curated case libraries (Saxena et al., 2014; Kim et al., 2024; Koubâa et al., 2024). Operationally, cloud-robotics frameworks and Robot Operating System (ROS)-level integrations such as ROSGPT provide software infrastructure for query-response reasoning between on-robot processes and cloud services (You et al., 2023; Koubâa et al., 2024). Edge-cloud coordination patterns offload heavy semantic reasoning to cloud while retaining real-time low-level control at edge, permitting robots to maintain safety-critical responsiveness while benefiting from expansive knowledge bases (Firoozi et al., 2024). Architectures like RoboEarth/Rapyuta operationalize this by pairing semantic encodings of actions and environments with remote computing generating grounded plans (Hunziker et al., 2013; Mohanarajah et al., 2015).

Large language models function as semantic mediators translating human intents into candidate plans, retrieving analogous precedents from case histories, or proposing task decompositions grounded by robot-specific affordance libraries (Kim et al., 2024; You et al., 2023). The literature stresses the need for grounding and affordance checks, that is the "do as I can, not as I say" constraint ensuring LLM-generated plans are validated against robot physical capabilities before execution (Ahn et al., 2022; Koubâa et al., 2024). Human roles shift toward strategic supervision, validation of candidate plans, and adjudication of risk or ethical boundaries moving from micromanagement toward oversight of consultative reasoning outputs (DeChant et al., 2023; Fang et al., 2025).

Foundational research on platforms and repositories established technical building blocks defining Level 5. RoboEarth and successors formalized languages and semantic representations for actions, objects, and environment models that robots query and ground during operation, demonstrating feasibility of a "web of robots" for executable knowledge sharing (Song et al., 2012; Hunziker et al., 2013; Singhal et al., 2017). Rapyuta operationalized integration of repository access with remote execution environments (Mohanarajah et al., 2015). RoboBrain and Open-EASE focus on scaling robot knowledge through extensive knowledge graphs and queryable explainability layers (Saxena et al., 2014; Beetz et al., 2015).

Natural language/LLM integrations illustrate the Level 5 trajectory. RoboGPT demonstrated how LLMs generate assembly sequences for construction tasks and reduce barriers for non-expert operators by offering semantically structured plans (You et al., 2023). Work integrating ChatGPT with robot middleware shows how on-robot processes call cloud LLM reasoning services, receive candidate plans, and map those into executable skills (Koubâa et al., 2024). LM-Nav provides instantiations where language, visual perception, and learned action priors combine to generate trajectories assessed and ranked prior to execution (Shah et al., 2022). Multi-layer LLM approaches to task decomposition show architectures propose hierarchical plans refined iteratively into lower-level actions when coupled with affordance checks (Kim et al., 2024; Luan et al., 2024).

Construction-oriented cloud platforms evidence the vertical axis: cloud BIM transmission and monitoring research document how building models, sensor feeds, and schedule/quality metadata can be centrally stored and queried by distributed agents (Afsari et al., 2016; Peng et al., 2017). DeChant et al. demonstrated how LLM-centered systems summarize robot prior actions and answer natural-language questions about past behavior, functionalities translating into case-history retrieval and post-hoc justification (DeChant et al., 2023). Multi-agent formulations using shared policies or federated updates enable robot fleets to exchange adaptations emerging at one site with peers at another, compressing experiential learning times (Kim et al., 2024; Firoozi et al., 2024). Lifelong federated reinforcement learning and federated imitation learning demonstrate robots enhance sample efficiency by transferring distilled policy knowledge across communities of agents and tasks (Liu et al., 2019, 2020). Semantic representation research shows explicit, standardized concept representations promote interoperability and reasoning over action affordances, object characteristics, and task breakdowns (Tenorth and Beetz, 2013; Militano et al., 2023).

In Level 5, Figure 7 and Table 5 show that cloud robotics, semantic knowledge bases, federated/foundation models, and language-based interfaces jointly pull the radar toward high Learning, Cognitive, and Improvisation scores. Representative clusters (RepIDs in Table T4) allow robots to query shared repositories, consult powerful cloud services, and use LLMs and language-to-action planners to synthesize candidate plans from large knowledge corpora and cross-project experience.

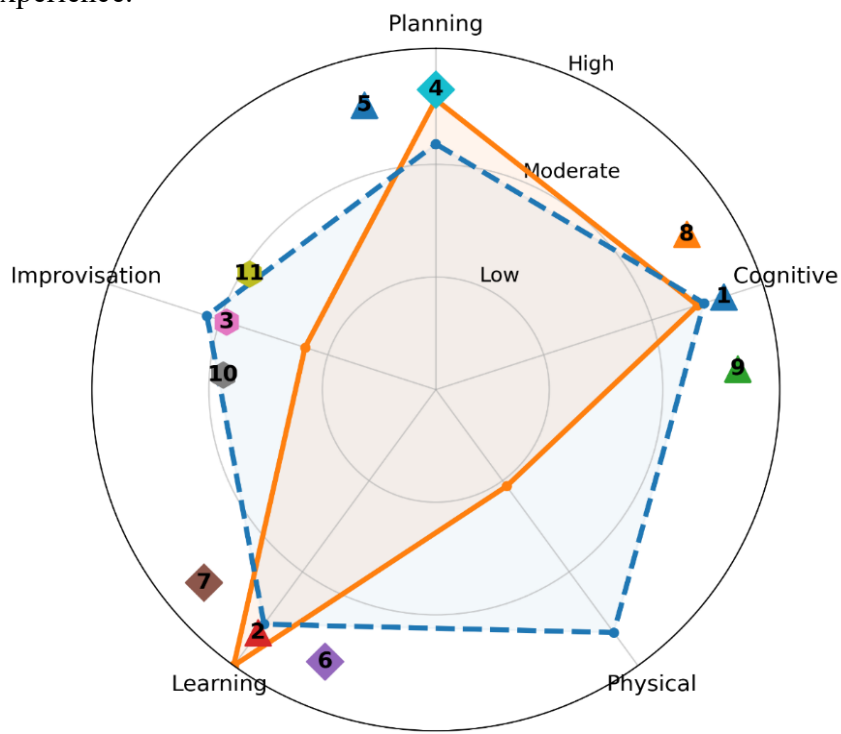

*Fig 7: Level 5: Cloud-Based Knowledge Systems – Representative Research Spoke-map*

*Table 5: Level 5 – Cloud-Based Knowledge: Representative Research Summary*

| RepID | Research Focus | Key Citation (Cluster) | Axis (primary skill advanced) | Shape (Family) | One-line Key Contribution |
|---|---|---|---|---|---|
| 1 | Cloud Robotics Platforms and Architectures | Kehoe et al. (2015); Wan et al. (2016); Hunziker et al. (2013); Mohanarajah et al. (2015) | Cognitive | ▲ | Cloud robotics platforms and middleware offload heavy computation and coordination, expanding robot cognitive capacity via shared cloud services. |
| 2 | Knowledge Repositories and Semantic Sharing | Song et al. (2012); Saxena et al. (2014); Beetz et al. (2015); Tenorth and Beetz (2013); Singhal et al. (2017) | Learning | ▲ | Semantic knowledge repositories let robots query skills, task models, and case histories, strengthening learning and cross-project generalization. |
| 3 | LLM-Robot Integration and Natural Language Interfaces | You et al. (2023); Koubâa et al. (2024); Kim et al. (2024); Luan et al. (2024) | Improvisation | ⬟ | LLM-based interfaces translate natural language into candidate plans and explanations, boosting cognitive reasoning and improvisation within safety bounds. |
| 4 | Language-to-Action and Multimodal Planning | Shah et al. (2022); Ahn et al. (2022); Yoshikawa et al. (2023); Mahadevan et al. (2024) | Planning | ◆ | Language-to-action and multimodal planners map instructions and perception into executable plans, raising robot planning capacity for known scenarios. |
| 5 | Cloud BIM and Construction-Specific Platforms | Afsari et al. (2016); Peng et al. (2017) | Planning | ▲ | Construction cloud BIM platforms centralize models and telemetry, supporting vertical consultation with remote experts who refine robot and team plans. |
| 6 | Federated and Lifelong Learning | Liu et al. (2019, 2020) | Learning | ◆ | Federated and lifelong learning allow robot fleets to exchange policy updates across sites, accelerating learning without centralizing raw data. |
| 7 | Foundation Models for Robotics | Firoozi et al. (2024) | Learning | ◆ | Foundation models for robotics provide pretrained representations and policies that can be adapted to new tasks, further boosting Level-5 learning potential. |
| 8 | Edge-Cloud Coordination and Middleware | Tenorth et al. (2012); Botta et al. (2021); Penmetcha et al. (2019); Ichnowski et al. (2022) | Cognitive | ▲ | Edge–cloud coordination frameworks balance low-latency on-robot control with cloud reasoning, enabling safe consultation of powerful remote services. |
| 9 | Semantic Representation and Ontologies | Tenorth and Beetz (2013); Militano et al. (2023) | Cognitive | ▲ | Semantic representations and ontologies support interoperable reasoning over objects, actions, and tasks, improving how Level-5 systems index knowledge. |
| 10 | Human Trust and Validation | DeChant et al. (2023); Ye et al. (2023); Fang et al. (2025); Robinson et al. (2025) | Improvisation | ⬟ | Trust and validation studies examine how humans calibrate reliance on cloud-augmented robots, ensuring that improvisation remains under human oversight. |
| 11 | Explainability and Shared Situational Awareness | Korpan and Epstein (2021); Lakhnati et al. (2024) | Improvisation | ⬟ | Explainability and shared-situation-awareness tools help robots summarize their reasoning and plans so humans can approve, modify, or override responses. |

*Improvisation Assessment for Level 5*

Level 5 systems exhibit high improvisatory competence but remain predominantly derivative and consultative rather than co-creative. We provisionally assess Level 5 improvisation capability at approximately 70%: robots handle wide range of unexpected situations by retrieving analogous precedents, recombining known action fragments, and proposing ranked alternatives, but do not yet engage in open, dialogic, co-creative ideation with human partners in real-time (Firoozi et al., 2024; Kim et al., 2024; Koubâa et al., 2024). This assessment is grounded in documented increases in functionality and generalization from foundation models and cloud sharing, tempered by limits of grounding, embodiment, and online generativity (Shah et al., 2022).

Level 5 improvisation manifests as intelligent retrieval, analogy, and recombination rather than spontaneous originality. Systems consult cloud archives and LLM knowledge to produce candidate courses of action derived from prior cases; outputs can be ranked, simulated, and constrained before execution (DeChant et al., 2023; Shah et al., 2022). Robots encountering unfamiliar assembly tolerances query shared BIM and case archives for similar tolerance mitigations, ask LLMs for decompositions, map candidates into robot-specific primitives, and present validated options to human supervisors (You et al., 2023).

However, several limitations prevent genuine co-creativity. First, LLM-centered reasoning remains insufficiently grounded in real-time perceptual affordances; natural language plans must be checked against robots' embodied capabilities (Ahn et al., 2022; Koubâa et al., 2024). Second, outputs are typically sequential query-response interactions rather than sustained dialogic cycles where humans and robots iteratively co-invent strategies through shared mental models (Yoshikawa et al., 2023). Third, the generativity gap which is the ability to propose genuinely novel structural strategies having no analogue in corpus, remains substantially unclosed, as most improvisation depends on recombination of known elements rather than invention of unprecedented tactics (Firoozi et al., 2024; Mahadevan et al., 2024).

Human workers continue to exercise final decision authority, particularly when ambiguity, moral judgment, or unprecedented structural risk is involved, while robots serve as high-throughput analytical and retrieval engines democratizing expertise (Fang et al., 2025; Robinson et al., 2025). Cloud-connected agents narrow the novice-expert gap by surfacing contextually relevant precedents and providing ranked alternatives scaffolding human judgment, while humans retain veto power and ethical oversight (Ye et al., 2023).

Level 5 represents culmination of data-driven, distributed intelligence integrating cross-project knowledge stores, cloud-hosted reasoning primitives, and horizontal/vertical consultation patterns materially expanding robots' repertoire for handling unforeseen conditions (Afsari et al., 2016; Firoozi et al., 2024). Critical elements separating Level 5 from Level 6 are emergence of genuinely shared mental models, dialogic co-reasoning updating both human and robot models in real time, and generative capacities permitting teams to invent novel strategies together, requiring advances in shared situational awareness interfaces, explainable architectures, and generative grounded reasoning systems (Yoshikawa et al., 2023; Lakhnati et al., 2024; Mahadevan et al., 2024; Firoozi et al., 2024).

**Level 6: True Collaborative Improvisation (Target State)**

*Target Capabilities and Representative Research*

Level 6 represents the conceptual pinnacle of the HRC maturity spectrum: an operational regime where human and robotic agents function as epistemic and creative partners, engaging in continuous, co-creative problem framing, exploration, and execution in real time (Hautala and Jauhiainen, 2022; Han et al., 2021). This level signifies a transition beyond knowledge-assisted, cloud-oriented inference toward co-creative cognition, wherein generative models, on-site digital twins, and edge/cloud knowledge infrastructures are integrated with embodied actuators and human sense-making, enabling emergent solution generation amid open-world uncertainty (Chen et al., 2024; Onososen and Musonda, 2022).

Level 6 is characterized by four interdependent capabilities yielding "improvisational parity." First, shared mental models demand persistent, dynamically aligned representational structures enabling partners to maintain synchronized task and context semantics across modalities and time horizons, operationalized for mutual prediction and expectation management (Talamadupula et al., 2014; Umbrico et al., 2024; Favier et al., 2023). Second, collaborative reasoning loops involve iterative cycles of perception, hypothesis, proposal, critique, and enactment, allowing both agents to initiate, evaluate, and revise plans through moment-to-moment closed-loop co-reasoning producing mutual improvement rather than linear handoffs (Musić and Hirche, 2017; Shafti et al., 2020; Ma et al., 2022). Third, generative, context-aware AI enables robots to generate candidate solutions, not only retrieve existing ones—that are contextually grounded in construction scenes, constraints, materials, and human preferences through integration of generative models with digital twins and on-site sensing (Chen et al., 2024; Jiang et al., 2019; Salzmann et al., 2020). Fourth, explainable autonomy and calibrated transparency ensure robot initiatives are interpretable and their rationale communicable in human-comprehensible terms, with explanations situated, actionable, and sensitive to interlocutors' epistemic states to support shared decision-making and maintain calibrated trust (Fisac et al., 2019; Stange and Kopp, 2023; Olivares-Alarcos et al., 2023).

Within this ensemble, human partners retain primacy for high-level contextual wisdom, ethical judgment, and strategic direction, while robotic partners supply high-bandwidth computational reasoning, predictive modeling, and rapid generative synthesis of alternative physical strategies (Han et al., 2021; Schaefer et al., 2020). Together, these capabilities enable joint decision-making in unpredictable, dynamic construction settings where neither an unaided human nor standalone robot can reliably produce optimal outcomes due to scale, speed, or multi-modal uncertainty (Drew, 2021).

Movement toward Level 6 is visible across convergent research trajectories—LLM and generative model integration, explainable AI and value alignment, multi-agent co-reasoning, and mixed-reality mediated interaction. Systematic reviews illustrate shifts from simple task automation toward integrated HRC paradigms emphasizing proximity, shared control, and knowledge exchange, while exposing substantial gaps in co-creative autonomy and on-site adaptability (Onososen and Musonda, 2022; Wei et al., 2023). Recent work demonstrates LLMs combined with digital twin technology in robot assembly, illustrating feasibility of robots proposing contextually aware, semantically rich plans that can be interrogated and adapted by human partners in real time (Chen et al., 2024). Transformer-based reinforcement learning, and generative planning research shows how sequence modeling can be applied to multi-robot exploration and decision generation, suggesting pathways for proposing context-sensitive alternatives in construction tasks (Schaefer et al., 2020; Ma et al., 2022). Open-world reasoning frameworks demonstrate how robots can expand knowledge representations dynamically, a prerequisite for credible on-the-fly proposal generation (Jiang et al., 2019).

Foundational work on explanation-need detection and pedagogic alignment frames Explainable AI (XAI) as a social process, providing algorithmic desiderata and evaluation dimensions for systems where robot explanations must be tailored to meet informational needs of human partners (Stange and Kopp, 2023; Fisac et al., 2019). Research on robot explanatory narratives confirms that socio-technical acceptance and transparent interactions are crucial for advancing shared autonomy, yielding design patterns for communicative behaviors and trust measurement instruments (Olivares-Alarcos et al., 2023). Advances in multi-agent reinforcement learning and human-aware planning illustrate methods for equipping agents with nested models of partners' intentions and coordinating divergent objectives toward joint solutions, supporting adaptive role negotiation and emergent division of labor (Ma et al., 2022; Shafti et al., 2020). Edge/cloud knowledge inference and semantic knowledge extraction frameworks allow low-latency, auditable knowledge sharing between multiple robots and human teams, facilitating distributed reasoning and trust calibration (Li et al., 2021; Shi et al., 2022).

Studies of virtual and augmented reality mediation reveal that multi-modal channels substantially improve situational awareness and coordination, enabling more fluid exchanges of context and intent than verbal commands alone (Xu et al., 2013; Lee et al., 2020). Research in augmented reality for timber prefabrication and masonry assembly illustrates how interactive displays and real-time visual feedback bridge human intent and robotic execution, enhancing on-site negotiation of alternatives (Böhme and Valenzuela-Astudillo, 2023; Song et al., 2023). Wearable haptics and embodied communication frameworks enhance physical cooperation and tacit signaling essential for improvisational teamwork (Musić et al., 2019; Peternel et al., 2017). Research combining embodied perception, trajectory forecasting, and sensorimotor coupling demonstrates how robots can anticipate human motion and make proactive adjustments necessary for safe, creative co-execution of tasks requiring tight physical coordination (Salzmann et al., 2020; Latella et al., 2018). However, benchmarking reveals that current systems typically exhibit sequential or human-dominant interaction paradigms rather than true co-creative parity, with implementations requiring human primacy in ideation or generating proposals lacking sufficient contextual grounding (Musić and Hirche, 2017; Raptis et al., 2025).

Finally, Figure 8 and Table 6 illustrate the research directions needed to approach true collaborative improvisation, where human and robot are both modeled at high capacity on all five axes but remain interdependent. Shared mental models, cognitive architectures, multi-agent co-reasoning, and generative/LLM-based methods (RepIDs in Table 6) aim to give robots the ability to propose genuinely new strategies and negotiate plans, while LfD/co-adaptation and control-sharing work focus on building joint routines and fluid exchange of initiative in physical tasks. Complementary streams on trust, value alignment, social robotics, and psychology anchor this vision in human oversight, emphasizing that even at Level 6, human collaborators are still needed to shape goals, resolve conflicts, and constrain improvisation to acceptable, context-specific behavior.

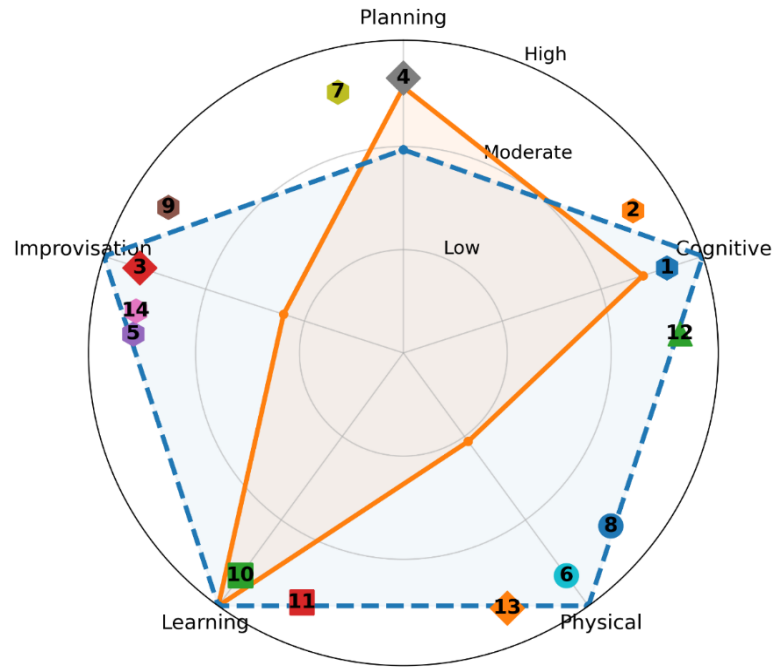

***Fig 8:*** *Level 6: True Collaborative Improvisation – Representative Research Spoke-map*

***Improvisation Assessment for Level 6***

At Level 6, improvisation shifts from an individual or assistive behavior to a co-evolutionary competence where humans and robots co-construct task representations, propose contingencies, and adapt strategies in reciprocal loops modifying both partners' capabilities (Hautala and Jauhiainen, 2022; Papadopoulos et al., 2021). This reframing treats improvisation as reactive that is responding to uncertainty and generative which is creating novel strategies, where novelty is evaluated and refined collaboratively (Zoelen et al., 2021; Ehresmann et al., 2023). Improvisation becomes creative search where generative models propose solution families and human partners refine proposals using contextual knowledge and ethical constraints, though grounding and verification remain critical to prevent unsafe recommendations (Raptis et al., 2025; Chen et al., 2024).

*Table 6: True Collaborative Improvisation: Representative Research Summary*

| RepID | Research Focus | Key Citation (Cluster) | Axis (primary skill) | Shape (Family) | One-line Key Contribution |
|---|---|---|---|---|---|
| 1 | Foundational Reviews and HRC Paradigms | (Han et al., 2021; Onososen & Musonda, 2022; Wei et al., 2023; Hautala & Jauhiainen, 2022) | Cognitive | ⬟ | Synthesizes trajectories toward co-creative HRC, clarifying gaps and requirements for improvisational parity in construction. |
| 2 | Shared Mental Models and Theory of Mind | (Talamadupula et al., 2014; Favier et al., 2023; Umbrico et al., 2024) | Cognitive | ⬟ | Develops shared mental models and theory-of-mind mechanisms so humans and robots can predict, coordinate, and realign plans jointly. |
| 3 | LLM and Generative AI for Robotics | (Chen et al., 2024; Raptis et al., 2025; Jiang et al., 2019; Schaefer et al., 2020) | Improvisation | ◆ | Integrates LLMs and generative models with robots and digital twins to propose novel, context-aware strategies during collaboration. |
| 4 | Multi-Agent Co-Reasoning and Coordination | (Ma et al., 2022; Shafti et al., 2020) | Planning | ◆ | Multi-agent RL and co-reasoning coordinate human and robot intentions for jointly optimized plans and role negotiation. |
| 5 | Explainable AI and Value Alignment | (Fisac et al., 2019; Stange & Kopp, 2023; Olivares-Alarcos et al., 2023; Robert, 2021) | Improvisation | ⬟ | XAI and value-alignment methods tailor explanations and objectives to human needs, supporting safe, value-consistent improvisation. |
| 6 | Control Sharing and Dynamic Autonomy | (Musić & Hirche, 2017; Musić et al., 2019; Schlossman et al., 2019; Peternel et al., 2017) | Physical | ● | Dynamic autonomy and control-sharing schemes let human and robot fluidly exchange initiative during tightly coordinated physical work. |
| 7 | Mixed Reality and Embodied Interfaces | (Böhme & Valenzuela-Astudillo, 2023; Song et al., 2023; Wang et al., 2024; Xu et al., 2013; Lee et al., 2020) | Planning | ⬟ | Mixed reality and embodied interfaces improve shared situational awareness and allow joint exploration of alternatives in context. |
| 8 | Embodied Perception and Trajectory Forecasting | (Salzmann et al., 2020; Latella et al., 2018) | Physical | ● | Trajectory forecasting and embodied perception enable robots to anticipate human motion and adjust proactively for safe co-execution. |
| 9 | Trust Inference and Calibration | (Guo et al., 2023; Li et al., 2021; Shi et al., 2021) | Improvisation | ⬟ | Trust inference and calibration models regulate when and how robot proposals are accepted, |
| | | | | | keeping authority distribution adaptive and safe. |
| 10 | Tacit Knowledge and Embodied Automation | (Fast-Berglund et al., 2018; Xu et al., 2022; Dinur, 2011) | Learning | ■ | Embodied automation and tacit-knowledge capture transfer craft expertise into forms that robots can learn and refine through co-execution. |
| 11 | Learning from Demonstration and Co-Adaptation | (Zoelen et al., 2021; Rozo et al. 2013; Wang et al. 2022) | Learning | ■ | LfD and co-adaptation studies show how repeated joint execution creates shared routines and mutual adjustments at team level. |
| 12 | Cognitive AI Architectures | (Papadopoulos et al., 2021; Ehresmann et al., 2023) | Cognitive | ▲ | Open cognitive architectures integrate reasoning, perception, and memory for scalable, explainable co-creative HRC in multi-agent settings. |
| 13 | Cross-Domain Applications | (Zhang & Tao, 2023; Drew, 2021) | Physical | ◆ | Cross-domain RL applications (e.g., sports, search and rescue) demonstrate high-performance coordination that can inform construction HRC. |
| 14 | Social Robotics and Human Psychology | (Bütepage & Kragić, 2017; Salem et al., 2009) | Improvisation | ⬟ | Social robotics and psychological studies characterize attitudes, gestures, and social cues essential for fluid, co-creative teamwork. |

A central metric is fluid authority distribution: control shifts dynamically between human and robot based on situational complexity and shared confidence estimates (Musić and Hirche, 2017; Schlossman et al., 2019). Human improvisation in construction relies on tacit knowledge, pattern recognition, and embodied judgment; translating these traits into robot-compatible forms depends on embodied automation approaches, tacit knowledge capture, and co-training paradigms (Fast-Berglund et al., 2018; Xu et al., 2022; Dinur, 2011). A comprehensive assessment framework should evaluate shared situational awareness, generativity, adaptive control dynamics, explainability and trust, and learning transfer (Umbrico et al., 2024; Olivares-Alarcos et al., 2023; Robert, 2021). Improvisation becomes collective competency learned at team level through repeated co-execution, establishing coordinated responses and shared heuristics (Hautala and Jauhiainen, 2022; Papadopoulos et al., 2021).

Substantial gaps remain: ethical alignment frameworks operationalizing values in generative models are under-development (Fisac et al., 2019; Papadopoulos et al., 2021); trust calibration requires robust methods preventing brittleness or over-reliance (Guo et al., 2023); encoding tacit craft knowledge remains an open challenge (Stange and Kopp, 2023); current deployments face practical constraints like sensing occlusion and unstructured terrains (Onososen and Musonda, 2022; Wei et al., 2023); and computational models of human intuition and ethical judgments are still developing (Ehresmann et al., 2023). Level 6 represents both conceptual peak of HRC trajectories and practical frontier requiring integrative breakthroughs in embodied cognition, generative models, shared knowledge representations, and methodologies making robot actions ethically clear (Raptis et al., 2025; Bütepage and Kragić, 2017). If realized, Level 6 would transform robots from high-performance assistants into creative collaborators, enabling teams to address construction unpredictability with adaptability and co-generated innovation that neither partner could achieve alone (Han et al., 2021; Hautala and Jauhiainen, 2022).

The comprehensive analysis of current capabilities across all six collaboration levels reveals critical patterns in technological advancement and persistent barriers that warrant deeper examination to guide future research directions.

**Critical Analysis and Outstanding Research Gaps**

The systematic review of improvisation capabilities across six levels of human-robot collaboration reveals a fundamental paradox: while technological advances have progressively enhanced robots' adaptive manipulation, learning, and knowledge consultation capabilities

(Levels 1-5), the transition to true collaborative improvisation (Level 6) remains obstructed by persistent technical, conceptual, and institutional barriers that existing research trajectories have yet to adequately address.

Technically, current systems exhibit three critical limitations. First, the grounding problem persists while Level 5 systems can retrieve analogous precedents from cloud knowledge bases and generate LLM-mediated plans, these outputs remain insufficiently grounded in real-time perceptual affordances and embodied capabilities, requiring explicit human validation before execution (Ahn et al., 2022; Koubâa et al., 2024). Second, interaction patterns remain predominantly sequential query-response cycles rather than sustained dialogic co-reasoning where humans and robots iteratively co-invent strategies through shared mental models (Yoshikawa et al., 2023; Firoozi et al., 2024). Third, the generativity gap which is the ability to propose genuinely novel structural strategies having no analogue in training corpus or knowledge repositories remains substantially unclosed, as improvisation depends primarily on recombination of known elements rather than invention of unprecedented tactics (Mahadevan et al., 2024).

Conceptually, the literature reveals a striking disconnect between human improvisation research in construction which emphasizes dyadic collaboration, vertical/horizontal consultation, and creative problem-solving under uncertainty (Hamzeh et al., 2019; Menches and Chen, 2013) and robotics research that predominantly focuses on individual robot capabilities rather than joint cognitive processes. This gap manifests in the absence of frameworks operationalizing how tacit craft knowledge, pattern recognition, and embodied judgment characteristic of human improvisation can be translated into robot-compatible representations supporting genuine co-creation (Fast-Berglund et al., 2018; Dinur, 2011). Furthermore, ethical alignment frameworks that operationalize values in generative models remain under-developed, raising unresolved questions about autonomous decisions in safety-critical construction contexts (Fisac et al., 2019; Papadopoulos et al., 2021).

Methodologically, the review identifies a critical research frontier between Level 5's consultative autonomy and Level 6's co-creative improvisation. Achieving this transition requires integrative breakthroughs across four domains: shared situational awareness interfaces enabling real-time model synchronization (Lakhnati et al., 2024); explainable and auditable architectures supporting transparent robot reasoning (Korpan and Epstein, 2021); generative, grounded reasoning systems coupling LLM-scale creativity with embodied affordance models

(Firoozi et al., 2024); and trust calibration mechanisms preventing brittleness or over-reliance in heterogeneous human-robot teams (Guo et al., 2023). Current research remains fragmented across these domains, lacking comprehensive frameworks integrating these elements into coherent Level 6 systems capable of transforming robots from high-performance assistants into creative collaborators addressing construction unpredictability through co-generated innovation.

**Future Research Directions**

Advancing from Level 5's consultative autonomy toward Level 6's true collaborative improvisation in construction requires coordinated research across three critical technological domains, each addressing specific barriers identified in this systematic review. First, augmented, and virtual reality (AR/VR) interfaces and explainable AI systems must enable real-time synchronization of spatial understanding, task representations, and constraint awareness between human workers and robotic systems (Lakhnati et al., 2024; Böhme and Valenzuela-Astudillo, 2023). Our research group is actively developing AR/VR interfaces that visualize robot reasoning processes, proposed actions, and confidence estimates in situ, enabling workers to rapidly validate or correct autonomous proposals during active construction sequences while supporting bidirectional information flow where human contextual knowledge updates robot models and robot computational analysis informs human decision-making (Umbrico et al., 2024; Stange and Kopp, 2023). Second, large language model (LLM) integration with robotic control systems requires hybrid architectures where generative models propose novel construction strategies such as alternative assembly sequences, adaptive material handling, creative problem decompositions, while perception-action loops validate feasibility through real-time sensing and physics-based simulation before execution, ensuring LLM-generated plans account for tacit craft knowledge, site-specific conditions, and safety boundaries characteristic of construction sites (Chen et al., 2024; Ahn et al., 2022; Park et al., 2024). Third, cloud-based knowledge systems must evolve beyond Level 5's sequential query-response patterns toward sustained dialogic co-reasoning where distributed knowledge bases support iterative strategy co-invention, requiring federated learning architectures enabling construction robot fleets to share improvisation experiences across projects while preserving proprietary knowledge, cloud-hosted case repositories encoding successful adaptations to material delays and geometric discrepancies, and semantic frameworks translating construction standards and best practices into machine-queryable formats accessible during real-time improvisation (Firoozi et al., 2024; You et al., 2023; Kim et al., 2024).

These research directions address the fundamental technical, conceptual, and methodological barriers identified in this review. Success requires demonstrating that integrated human-robot systems can jointly navigate construction's inherent unpredictability through co-creative problem-solving that leverages complementary human and robot strengths while preserving essential human roles in ethical judgment, contextual wisdom, and strategic decision-making. The proposed taxonomy and analysis framework provide clear pathways for positioning ongoing research contributions and identifying specific technical requirements as the construction industry progresses toward truly collaborative human-robot partnerships.

**Conclusions**

This research presents a comprehensive six-level taxonomy for classifying human-robot collaboration in construction based on improvisation capabilities, providing a systematic framework for understanding the evolution from manual execution through preprogramming, adaptive manipulation, imitation learning, and human-in-loop BIM integration toward cloud-based knowledge systems and ultimately true collaborative improvisation. The systematic review of 214 articles reveals that current construction robotics research concentrates predominantly at lower taxonomy levels.

The radar diagram analysis demonstrates progressive capability evolution across five critical dimensions that is Planning, Cognitive Role, Physical Execution, Learning Capability, and Improvisation, revealing that while robots increasingly surpass human physical capabilities and task-level cognitive reasoning at higher levels, true collaborative improvisation requires intentional asymmetries preserving essential human roles in high-level planning, contextual learning, and ethical judgment. The research identifies three fundamental barriers obstructing advancement toward Level 6 collaborative improvisation: technical limitations including insufficient grounding of LLM-generated plans in real-time perceptual affordances and predominantly sequential query-response interactions rather than sustained dialogic co-reasoning; conceptual gaps manifesting as disconnects between human improvisation research emphasizing dyadic collaboration and robotics research focusing on individual robot capabilities; and methodological challenges requiring integrative breakthroughs across shared situational awareness interfaces, explainable architectures, generative grounded reasoning systems, and trust calibration mechanisms.

Future research must prioritize development of AR/VR interfaces and explainable AI for shared mental models, integration of large language models for intuitive communication while ensuring robust grounding in construction-specific constraints, and advancement of cloud-based knowledge systems enabling horizontal and vertical consultation patterns mimicking human improvisational processes. The proposed taxonomy provides researchers and practitioners with a structured framework for positioning ongoing work along the maturity spectrum, identifying relevant comparative studies, and recognizing specific technical requirements and research gaps. By establishing clear pathways from current capabilities toward true collaborative improvisation while maintaining human expertise primacy, this framework guides the construction industry's evolution from viewing robots as mere tools toward embracing them as creative partners capable of jointly addressing construction unpredictability through co-generated innovation that neither humans nor robots could achieve independently.

**Data Availability Statement**

All data, models, and code generated or used during the study appear in the submitted article.

**Acknowledgments**

The authors would like to acknowledge the financial support for this research received from the US National Science Foundation (NSF) (Grant Nos. FW-HTF 2025805 and FW-HTF 2128623). Any opinions and findings in this paper are those of the authors and do not necessarily represent those of the NSF.